\documentclass{article}

\PassOptionsToPackage{table}{xcolor}
\usepackage{fullpage}
\usepackage[utf8]{inputenc}
\usepackage[T1]{fontenc}
\usepackage{url}
\usepackage{booktabs}
\usepackage{nicefrac}
\usepackage{microtype}
\usepackage{xcolor}
\usepackage{graphicx}
\usepackage{wrapfig}
\usepackage{subcaption}
\usepackage{enumitem}
\usepackage[section]{placeins}
\usepackage{amsmath}
\usepackage{amssymb}
\usepackage{amsfonts}
\usepackage{mathtools}
\usepackage{amsthm}
\usepackage{multirow}
\usepackage{colortbl}
\usepackage{siunitx}
\usepackage{adjustbox}
\usepackage[round,semicolon,sort]{natbib}
\definecolor{mydarkblue}{rgb}{0,0.08,0.45}
\usepackage[colorlinks,citecolor=mydarkblue,urlcolor=mydarkblue,linkcolor=mydarkblue]{hyperref}
\usepackage[capitalize,noabbrev]{cleveref}
\let\cite\citep

\usepackage{amsmath,amsfonts,bm}

\def\eqref#1{equation~\ref{#1}}

\def\1{\bm{1}}

\DeclareMathAlphabet{\mathsfit}{\encodingdefault}{\sfdefault}{m}{sl}
\SetMathAlphabet{\mathsfit}{bold}{\encodingdefault}{\sfdefault}{bx}{n}

\theoremstyle{plain}

\theoremstyle{definition}

\theoremstyle{remark}

\newcommand{\newupdate}{}

\title{AlphaQ: Calibration-Free Bit Allocation for \\ Mixture-of-Experts Quantization}
\date{}
\author{{\bf Wanqi Yang}\textsuperscript{\normalfont 1,2,3} \enspace
	{\bf Yuexiao Ma}\textsuperscript{\normalfont 4} \enspace
	{\bf Alexander Conzelmann}\textsuperscript{\normalfont 1,2} \enspace
	{\bf Xiawu Zheng}\textsuperscript{\normalfont 4} \\
	{\bf Michael W. Mahoney}\textsuperscript{5,6,7} \enspace
	{\bf T. Konstantin Rusch}\textsuperscript{1,2,3,8} \enspace
	{\bf Shiwei Liu}\textsuperscript{1,2,3} \\
	\textsuperscript{1}Max Planck Institute for Intelligent Systems \\
	\textsuperscript{2}ELLIS Institute Tübingen \enspace
	\textsuperscript{3}Tübingen AI Center \\
	\textsuperscript{4}Key Laboratory of Multimedia Trusted Perception  and Efficient Computing, Xiamen University \\
	\textsuperscript{5}International Computer Science Institute \enspace
	\textsuperscript{6}Lawrence Berkeley National Laboratory \\
	\textsuperscript{7}University of California, Berkeley \enspace
	\textsuperscript{8}Liquid AI
}

\begin{document}

	\maketitle
	
	\begin{abstract}
		Mixture-of-Experts (MoE) architectures scale model capacity through sparse expert activation, but their deployment remains memory-bound because all expert weights must reside in memory. Mixed-precision quantization can substantially reduce this footprint by assigning different bit-widths to different experts. Existing approaches, however, typically rely on calibration data to estimate expert importance and determine bit allocation. For frontier MoE LLMs, the original training data, and hence the true training distribution, is proprietary and inaccessible. As a result, calibration sets are inevitably imperfect surrogates, and this can misestimate expert utilization and lead to suboptimal bit allocation. Motivated by the substantial cross-expert quality variability observed in modern MoE models, and by the success of \emph{Heavy-Tailed Self-Regularization (HT-SR) theory} at predicting neural network model quality without access to training or testing data, we propose \textbf{AlphaQ}, a \emph{calibration-free bit-allocation method for MoE quantization}. AlphaQ draws on HT-SR theory and follows a simple principle: experts with more heavy-tailed weight spectra are typically better trained and hence should receive higher bit-widths, while experts with weaker heavy-tailed structure can be quantized more aggressively. AlphaQ operationalizes this principle by measuring expert-wise spectral heavy-tailedness and solving a budget-constrained optimization problem that minimizes total quantization error under a global bit-budget constraint. Across several MoE models, AlphaQ consistently outperforms calibration-based baselines under matched bit budgets. Notably, on Qwen1.5-MoE, AlphaQ achieves near full-precision accuracy with an average expert precision of only \textbf{3.5} bits, while delivering more than \textbf{4$\times$} memory compression. Our code is available at \href{https://github.com/Superone77/AlphaQ}{github.com/Superone77/AlphaQ}.

	\end{abstract}
	\section{Introduction} 
	
	Mixture-of-Experts (MoE)~\cite{jacobs1991adaptive,jordan1994hierarchical,shazeer2017outrageously,lepikhin2020gshard,fedus2022switch,zoph2022designing} have received widespread attention due to their computational efficiency: by routing each token to a small subset of experts, MoE achieves a strong trade-off between quality and efficiency at massive parameter counts. 
	However, this sparsity often does \emph{not} translate into memory reduction at deployment time. 
	During inference, all expert weights must remain resident in GPU memory, making storing the expert weights the primary memory bottleneck. 
	Quantization is therefore an attractive path to make MoE deployable at scale~\cite{frantar2023qmoe,li2024examining,hu2025moequant,tao2025moqae,zheng2025dynamo,chen2025eac}. However, since experts (and even layers within each expert) contribute differently to model performance~\cite{sun2025curse,hu2025moequant,MM20a_trends_NatComm,yang2023test}, precisely allocating bits across experts and layers remains challenging.
	

	To address this challenge, most existing MoE quantization methods adopt a data-driven pipeline and derive mixed-precision configurations from input-dependent statistics~\cite{huang2024mixture,duanmu2025mxmoe,anonymous2025towards,xie2025automated}.
	While effective in practice, these approaches suffer from two fundamental limitations: local optimality; and dependence on calibration data.

	First, some of these methods~\cite{huang2024mixture,duanmu2025mxmoe,xie2025automated} allocate bits \emph{locally} under a fixed budget, e.g., using the same bit budget for every block or expert, overlooking the fact that different Transformer blocks or experts contribute unequally to global performance. Such locally optimal choices can be globally suboptimal~\cite{anonymous2025towards}.
	
	Second, since the original training data for frontier MoE LLMs is usually proprietary and inaccessible, existing quantization methods have to rely on biased and incomplete \emph{calibration data} to estimate expert importance. Notably, an imperfect or non-representative calibration set may fail to activate certain experts, which yields biased importance estimates~\cite{zheng2025dynamo,hu2025moequant}. Consequently, calibration-based bit allocation directly couples the performance of quantized models to the calibration distribution, ultimately leading to poor cross-domain generalization.
	
\begin{figure}
		\centering
		\includegraphics[width=1\linewidth]{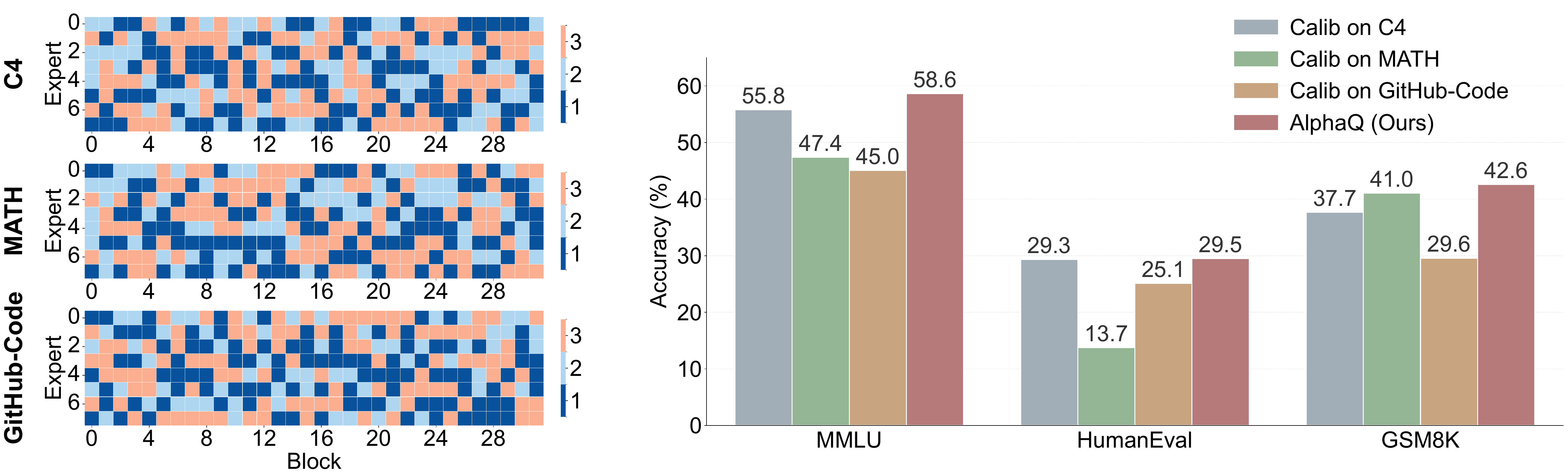}
		\caption{\textbf{Domain bias introduced by data-driven bit-width allocation in Mixtral-8$\times$7B.} Left: bit-width allocations calibrated on datasets across domains (C4~\cite{raffel2020exploring}, MATH~\cite{hendrycksmath2021}, GitHub-Code~\cite{swift_github_code_2024}) illustrate calibration-data-induced variations. Right: Mixtral-8$\times$7B calibrated on these datasets with a 2.5-bit budget exhibits performance bias, overfitting to the calibration domain and degrading on unseen tasks.  This indicates that data-driven bit allocation reduces cross-domain generalization by coupling performance to the calibration distribution.}
		\label{fig:calibration_bias_performance}
	\end{figure}

	\looseness=-1 Figure~\ref{fig:calibration_bias_performance} illustrates this effect: Mixtral-8$\times$7B~\cite{jiang2024mixtral}, quantized by the data-driven PMQ method~\cite{huang2024mixture} with different domain-specific datasets, exhibits substantially different mixed-precision allocations (expert activation patterns for different datasets are also provided in Appendix~\ref{app:activation_pattern}), leading to highly skewed performance across common-sense tasks (MMLU~\cite{hendryckstest2021}), math reasoning (GSM8K~\cite{gsm8k}), and coding (HumanEval~\cite{chen2021evaluating}).
	For instance, a model calibrated on MATH excels at math reasoning, but it lags behind GitHub-Code-calibrated models on coding tasks, indicating overfitting to the calibration domain and degraded robustness on unseen domains. While an increasing number of approaches aim to improve the quality of calibration statistics~\cite{zheng2025dynamo,hu2025moequant}, it remains fundamentally challenging for data-driven approaches to cover all potential domains.
	
	In this paper, we tackle this challenge by proposing \textbf{AlphaQ}, a novel calibration-free\footnote{We emphasize that the term \textit{calibration-free} applies only to the bit-allocation stage. The subsequent quantization step can be implemented with any existing quantization method~\cite{quantiz_review_TR21,GYKx24_memory_wall_TR}.} (and thus data-independent) bit-allocation method tailored for MoE quantization. 
	AlphaQ is inspired by \emph{Heavy-Tailed Self-Regularization (HT-SR) theory}~\cite{mahoney2019traditional}, which substantiates the principle that experts with heavier-tailed weight spectra are empirically linked to more informative structure and stronger correlations~\cite{MM20_SDM,MM20a_trends_NatComm,martin2021implicit}. 
	Concretely, AlphaQ characterizes each expert’s training sufficiency and importance from the shape of its weight-matrix spectrum, and it allocates bit-widths under a global precision budget: that is, experts with heavier-tailed eigenvalue spectra receive higher precision, whereas relatively under-trained or less sensitive experts with lighter-tailed spectra are quantized more aggressively.

	{\newupdate
		At the core of AlphaQ is a unified, data-agnostic importance metric for quantifying expert disparity across MoE architectures and layers. 
		Using this metric, we observe clear importance diversity across MoE families, within a given model, and even within a given block.
		Notably, fine-grained, highly sparse MoEs (e.g., DeepSeekV2-Lite~\cite{liu2024deepseek}, Qwen1.5-MoE~\cite{qwen_moe}) show larger variance, whereas vanilla MoEs with fewer experts (e.g., Mixtral-8$\times$7B) show much smaller variance.
	}
	
\begin{figure}
		\centering
		\includegraphics[width=1\linewidth]{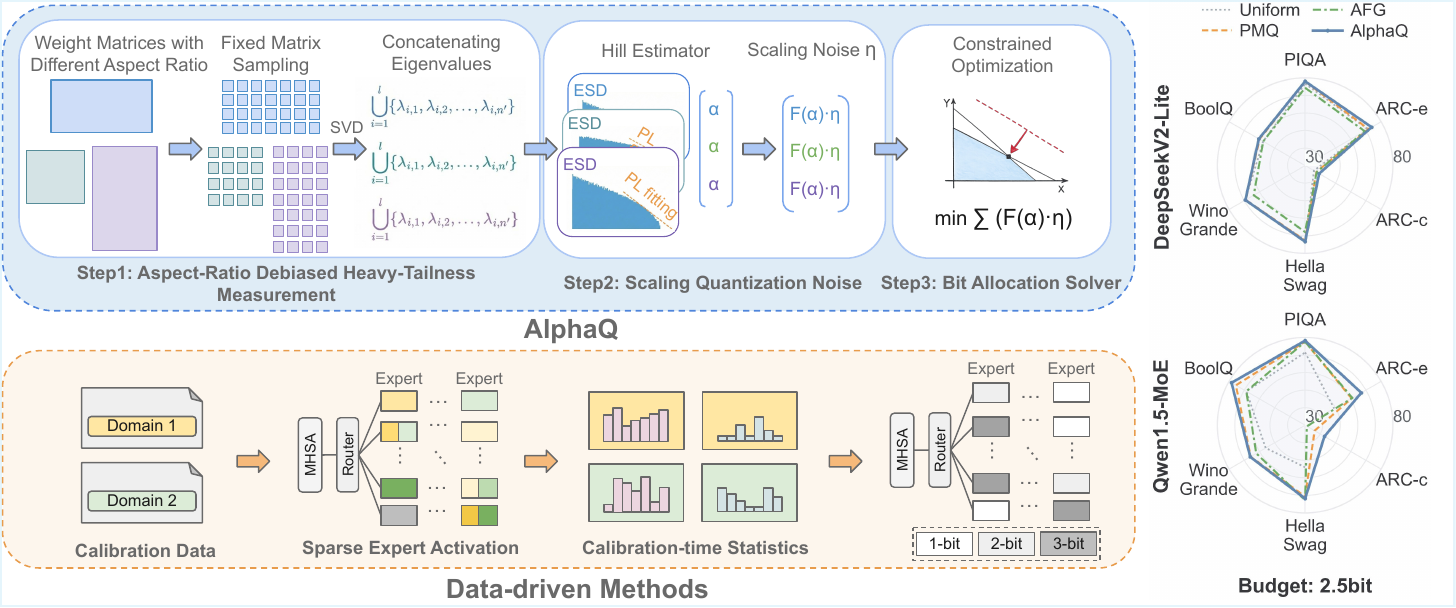}
		\caption{Comparison of the proposed (data-independent) AlphaQ framework and data-driven (or data-dependent) methods for MoE quantization.}
		\label{fig:workflow}
	\end{figure}

	
	{\newupdate
		Based on this calibration-free importance signal, AlphaQ formulates mixed-precision quantization as a budget-constrained optimization problem. Our formulation is flexible enough to accommodate the large importance variance observed across MoE architectures, across models, depths, and widths, as well as across submodules within each expert. \cref{fig:workflow} illustrates the comparison between AlphaQ and data-driven methods.
	}
	
	{\newupdate
		Overall, we demonstrate that, across a wide range of MoE-LLMs and benchmarks, AlphaQ reduces quantization-induced accuracy loss and outperforms existing calibration-based bit-allocation methods under matched bit budgets.
		With our quantized inference backend, the resulting low-bit MoE models further reduce the  weight memory footprint and improve end-to-end inference speed.
		Notably, with a 3.5-bit budget, Qwen1.5-MoE matches the BF16 model in accuracy while using just one-quarter of its weight memory footprint.
	}
	
	
	
	
	

	\section{Background and Related Work}
	
	\subsection{MoE Quantization}
	By representing weights and activations in low-precision formats, post-training quantization (PTQ)~\cite{dettmers2022gpt3,frantar2022gptq,xiao2023smoothquant,lin2024awq,liu2024spinquant} substantially reduces the storage and computational cost of LLMs. 
	Recently, PTQ for MoE-LLMs has received increasing attention. 
	Prior works~\cite{huang2024mixture,duanmu2025mxmoe,anonymous2025towards,xie2025automated} primarily formulate optimization problems to assign different quantization precisions to sparsely activated experts. 
	PMQ in MC-MoE~\cite{huang2024mixture} allocates precision by solving a transformer-block-wise linear programming problem using expert activation frequency and reconstruction error measured on calibration data. GEMQ~\cite{anonymous2025towards} extends this approach by performing precision allocation at a global level. MxMoE~\cite{duanmu2025mxmoe} uses expert activation frequency and sensitivity from calibration data to automatically generate customized kernels, enabling substantial speedups at comparable bit budgets. Furthermore, to account for the dependence of expert activation on calibration data, MoEQuant~\cite{hu2025moequant} proposes an expert-balanced sampling strategy for calibration, which improves traditional quantization methods in MoE but does not remove the limitation of calibration in bit allocation. 
	DynaMo~\cite{zheng2025dynamo} emphasizes that MoE quantization should be aware of data-model distribution shifts by modeling expert dynamics across different input distributions. However, their experiments mainly consider transfer across general-purpose text distributions (e.g., WikiText2, C4), rather than explicitly exposing domain-dependent calibration bias in bit allocation. By contrast, AlphaQ directly targets this failure mode and avoids it through calibration-free global bit allocation.
	
	\subsection{HT-SR Theory and Metrics}
	\label{HT-SR}

	\looseness=-1 Heavy-Tailed Self-Regularization (HT-SR) theory provides a principled framework to analyze neural network training sufficiency by examining the empirical spectral density (ESD) of layer-wise weight correlation matrices~\cite{mahoney2019traditional,MM20_SDM,MM20a_trends_NatComm,martin2021implicit}.
	Rooted in Random Matrix Theory~\cite{couillet2022random,models_htmu_TR} and the statistical mechanics of learning~\cite{EB01_BOOK,MM17_TR}, HT-SR posits that the heavy-tailed structure of trained weight matrices’ ESDs strongly predicts model performance~\cite{MM20a_trends_NatComm,yang2023test}.
	ESDs are often modeled with “spikes” (ground-truth-aligned learned features) and a “bulk” (noise following the Marchenko-Pastur law~\cite{wang2023spectral}), but HT-SR exploits the empirical fact that modern state-of-the-art networks are “strongly correlated systems,” making heavy-tailed spectral metrics more suitable.
	Informally, ESD heavy-tailed distributions stem from spike-bulk interaction, marking a critical “bulk-decay” phase for well-trained models~\cite{kothapalli2024crafting}.
	HT-SR metrics include “scale metrics” (correlating with generalization gaps~\cite{yang2023test}, but not with self-reported model quality) and “shape metrics,” e.g., fitted power-law exponents ($\alpha$)~\cite{MM20a_trends_NatComm} or the robust \texttt{PL\_Alpha\_Hill}~\cite{zhou2023temperature}. The latter capture weight matrix structural quality for model diagnostics and pruning~\cite{liu2024model,lu2024alphapruning,hu2025eigenspectrum,He2025AlphaDecay}, even predicting state-of-the-art model quality without training/testing data~\cite{MM20a_trends_NatComm,yang2023test,he2026alphadecay,he2026one}.

	{\newupdate
		\section{AlphaQ}
		In this section, we introduce \textbf{AlphaQ}. We first define our notation and compute our calibration-free importance score from model weights using \texttt{PL\_Alpha\_Hill}. We then formulate a global, budget-constrained optimization to assign mixed precisions across experts and submodules.
	}

	\subsection{Notation}
	
	In this work, we define a \emph{block} as a Transformer block. 
	In popular MoE models, each block comprises two modules: an attention module and an MoE module containing multiple experts and a routing layer. 
	Each attention or MoE module consists of multiple \emph{layers} (e.g., projection layers). A terminology summary is provided in Appendix~\ref{app:teminology_summary}.
	
	We consider a MoE model with $L$ layers. Let $\mathbf{W}_i \in \mathbb{R}^{m \times n}$ denote the weight matrix of the $i$-th layer, and define the corresponding correlation matrix as $\mathbf{X}_i = \mathbf{W}_i^\top \mathbf{W}_i$.
	The empirical spectral density (ESD) of $\mathbf{X}_i$, viewed as a probability measure over the eigenvalue distribution of $\mathbf{X}_i$, is defined as
	\begin{equation}
		\mu_{\mathbf{X}_i} := \frac{1}{n} \sum_{j=1}^{n} \delta_{\lambda_j(\mathbf{X}_i)},
		\label{eq:empirical_spectral_densit}
	\end{equation}
	where $\lambda_1(\mathbf{X}_i) \le \cdots \le \lambda_n(\mathbf{X}_i)$ are the eigenvalues of $\mathbf{X}_i$, and $\delta$ denotes the Dirac delta function.
	
	\subsection{Estimating Layer Importance in MoE}
	\label{sec:pl_alpha}
	
	\label{sec:Alpha_Score}
	When calibration-time activations are unreliable or biased, we derive the bit allocation signal directly from model parameters.
	We leverage the empirical observation that weight matrix spectral structure encodes learned correlations and noise sensitivity~\cite{mahoney2019traditional,MM20_SDM,lu2024alphapruning}.
	Specifically, HT-SR theory demonstrates that layers extracting more informative features develop heavy-tailed ESDs of their weight matrices, indicating better training and more effective inference signal extraction~\cite{lu2024alphapruning}.
	We therefore estimate layer importance based on ESD heavy-tailedness: weight matrices with strongly heavy-tailed ESDs encode richer correlation structures and receive higher bit-widths, while layers with lighter-tailed ESDs receive lower bit-widths.

	{\newupdate
		\textbf{PL\_Alpha\_Hill Metric}. 
		To quantify the heavy-tailed characteristics of weight matrix ESDs, we employ the \texttt{PL\_Alpha\_Hill} metric~\cite{zhou2023temperature}.\footnote{This \texttt{PL\_Alpha\_Hill} metric is ``biased,'' relative to the original HT-SR fitted power law $\alpha$ metric~\cite{MM20a_trends_NatComm}, but it correlates with $\alpha$ in the region of interest, and it is more robust to estimate.} 
		This metric partitions the eigenvalue domain into equal-width bins and counts the eigenvalues within each. We define the ESD tail as eigenvalues from the bin with maximum count along with all eigenvalues exceeding this bin's upper bound, which is then modeled using a power-law (PL) density:
		\begin{equation}
			p(\lambda) \propto \lambda^{-\alpha}, \qquad \lambda_{\min} < \lambda < \lambda_{\max},
		\end{equation}
		where the exponent $\alpha$ characterizes the degree of heavy-tailedness, with smaller values indicating heavier tails.
		We then estimate the exponent $\alpha$ via the Hill estimator~\cite{hill1975simple} on the ESD tail: 
		\begin{equation}
			\alpha
			= 1 + \left(\frac{1}{k}\sum_{i=1}^{k}\ln\frac{\lambda_{n-i+1}}{\lambda_{n-k}}\right)^{-1},
			\label{eq:pl_alpha_hill}
		\end{equation}
		\looseness=-1 where $\{\lambda_i\}_{i=1}^{n}$ are the eigenvalues used for tail fitting, sorted in ascending order, and $k$ determines the lower threshold $\lambda_{\min}$ for truncated power-law estimation.  
        We select $k$ using the \textit{Fix-finger} method~\cite{yang2023test}, which aligns $\lambda_{\min}$ with the peak of the ESD. 
		To overcome the effect of matrix aspect ratio on spectral shape measurements~\cite{hu2025eigenspectrum}, we use the \emph{Fixed-Aspect-Ratio Matrix Subsampling} (FARMS)~\cite{hu2025eigenspectrum}. 
        The derivation of Eq.~\ref{eq:pl_alpha_hill} is provided in Appendix~\ref{app:hill-derivation}. 
		Following HT-SR theory, we interpret smaller \texttt{PL\_Alpha\_Hill} values as stronger spectral heavy-tailedness and higher relative importance for bit allocation.
		Notably, \texttt{PL\_Alpha\_Hill} is a calibration-free metric, i.e., it does not require access to training or testing data, and it is a globally-comparable importance metric.
	}
	{\newupdate
		\begin{figure}[t]
			\centering
			\includegraphics[width=1\linewidth]{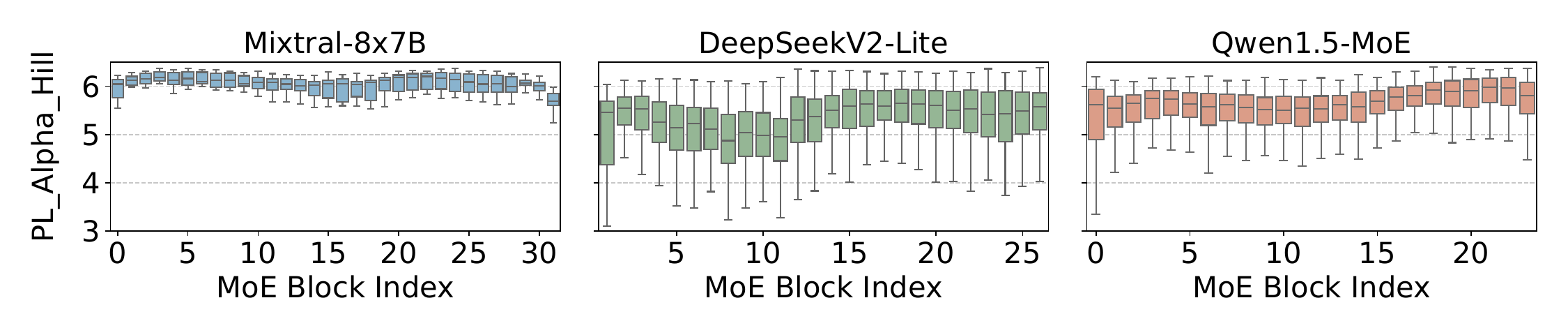}
			\caption{\textbf{Distribution of \texttt{PL\_Alpha\_Hill} across all up/gate/down projections in three representative MoE-LLMs.} The bottom and top of each boxplot indicate the minimum and maximum values of \texttt{PL\_Alpha\_Hill} across all up/gate/down projections within the block. The lower and upper edges of the box correspond to the first and third quartiles for that block, respectively, and the horizontal line inside the box denotes the median value.}
			\label{fig:alpha_boxplot_overview}
		\end{figure}
	}
    {\newupdate
		\begin{figure}[t]
			\centering
			\includegraphics[width=1\linewidth]{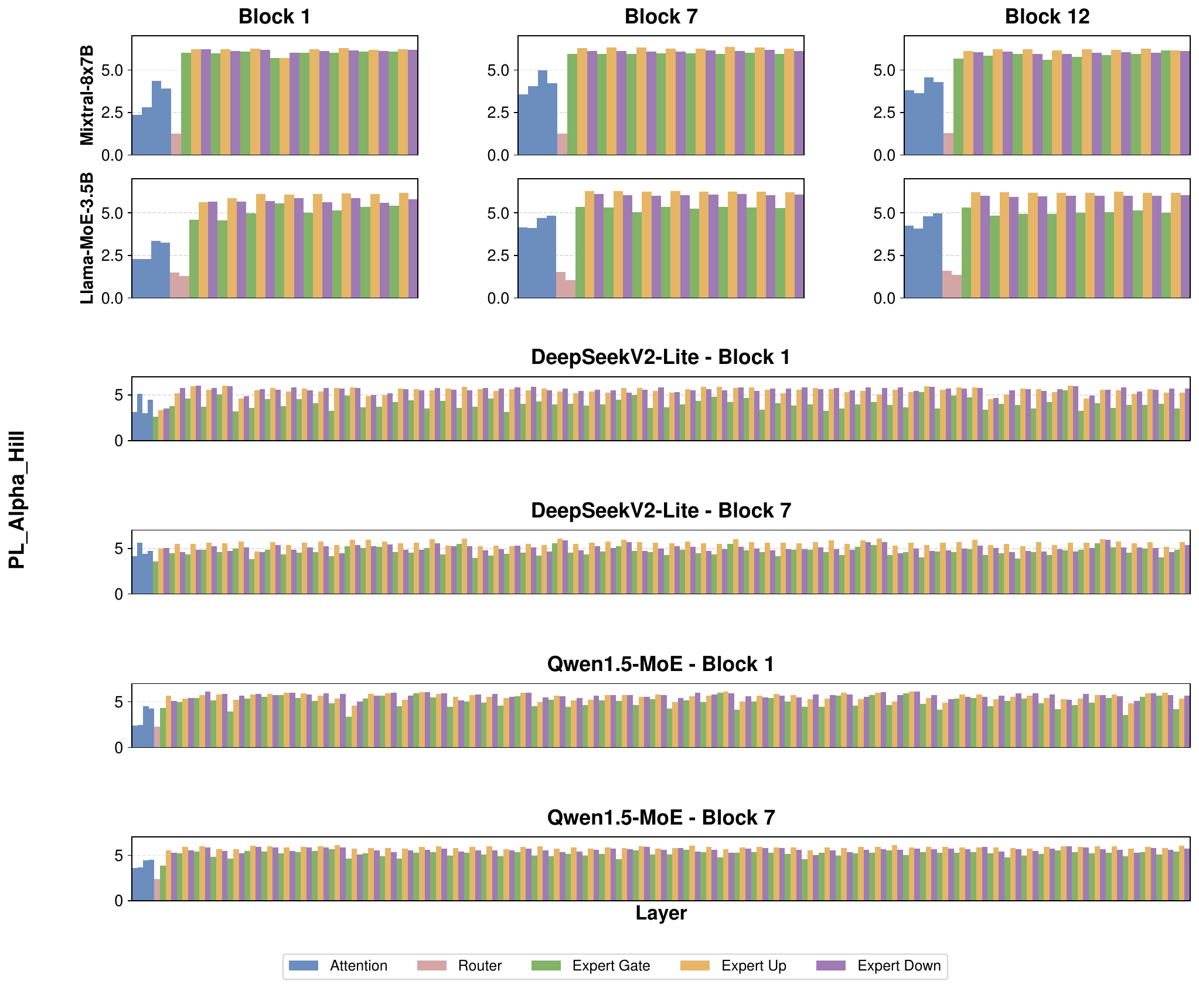}
			\caption{\textbf{Layer-wise \texttt{PL\_Alpha\_Hill} distribution in sampled MoE blocks.} The up, gate, and down projections within the same MoE block often have different \texttt{PL\_Alpha\_Hill} values, motivating layer-wise rather than expert-wise bit allocation.}
			\label{fig:moe_alpha_barchart}
		\end{figure}
	}
    
	{\newupdate
		\noindent\textbf{Alpha-based Importance Analysis.}
		We analyze popular MoE models using \texttt{PL\_Alpha\_Hill}. As shown in~\cref{fig:alpha_boxplot_overview}, \texttt{PL\_Alpha\_Hill} exhibits clear variation across layers, blocks, and architectures.
		
		\textit{i) Layer- and expert-level heterogeneity.}
		Within the same MoE block, different projection layers exhibit distinct \texttt{PL\_Alpha\_Hill} values, as further illustrated in~\cref{fig:moe_alpha_barchart}, showing that assigning a single bit-width to an entire expert can miss layer-level differences. Furthermore, expert-wise \texttt{PL\_Alpha\_Hill} variation within the same MoE layer suggests that uniform expert precision can under-protect important experts. Additional expert-wise evidence is provided in Appendix~\ref{app:more_detailed_alpha}.
		
		\textit{ii) Block-level heterogeneity.}
		Different MoE blocks have distinct \texttt{PL\_Alpha\_Hill} distributions, suggesting that a fixed block-wise budget can be suboptimal. Instead, the bit budget should be allocated globally across blocks so that more important regions of the model receive higher precision.
		
		\textit{iii) Architecture-level heterogeneity.}
		The distribution of \texttt{PL\_Alpha\_Hill} also differs across MoE architectures. Vanilla MoEs such as Mixtral-8$\times$7B show relatively small within-block variance, whereas fine-grained MoEs such as DeepSeekV2-Lite and Qwen1.5-MoE show much larger variance. This suggests that the mapping from \texttt{PL\_Alpha\_Hill} to importance should adapt to the model-level \texttt{PL\_Alpha\_Hill} distribution.
		
		These observations motivate our design: bit allocation should use the layer-wise  \texttt{PL\_Alpha\_Hill} distribution to decide where the global bit budget should be spent. We therefore formulate AlphaQ as a budget-constrained optimization problem that converts our calibration-free importance signal into layer-wise bit assignments.
	}

	{\newupdate
		
		\subsection{Bit Allocation Optimization}
		\label{sec:bit_allocation_opt}
		After establishing a globally comparable importance signal, we next allocate bit-widths to minimize accuracy degradation under a fixed global bit budget.
	}
	
	{\newupdate
		\noindent\textbf{Importance-Scaled Quantization Noise.}
		Using an importance metric alone is insufficient to determine the bit-widths: bit allocation requires quantifying both the benefit of assigning higher precision and the cost of reducing precision elsewhere in the budget.
		While \texttt{PL\_Alpha\_Hill} provides a calibration-free importance estimate, it does not capture the numerical perturbation induced by low-bit quantization.
		Conversely, quantization-induced numerical perturbation (i.e.,\ quantization noise) alone does not indicate whether the affected module is structurally important. Appendix~\ref{app:alpha_noise_analysis} provides a detailed analysis of this interaction.
	}
	
	{\newupdate
		We model the magnitude of quantization noise as a function of both the quantization bit-width and the weight distribution. Under the standard uniform-noise approximation, the quantization noise is modeled as zero-mean noise whose variance is proportional to the squared quantization step size. Since the step size decays exponentially with the bit-width $b$, i.e., $\Delta_{l,b} \propto 2^{-b}$, the corresponding noise variance scales as $2^{-2b}$. The derivation is provided in Appendix~\ref{app:quant_noise_model}.
		To account for layer-wise differences in weight distribution, we modulate this term by the variance of the weight matrix $\mathrm{Var}(\mathbf{W}_l)$ and use $\mathrm{Var}(\mathbf{W}_l)2^{-2b}$ to quantify layer-wise quantization noise.
		
		We then scale this quantization noise by alpha-based importance, penalizing perturbations more strongly on important layers. Specifically, we scale layer-wise quantization noise by $\left( \tilde{\alpha}/{{\alpha}_l} \right)^{\gamma}$, where $\tilde{\alpha} = \operatorname{median}\{\alpha_l\}_{l=1}^{L}$ and $\gamma$ is a data-free curvature parameter. Let $\alpha_{\min}$, $\alpha_{\max}$, and $v_\alpha$ denote the minimum, maximum, and variance of \texttt{PL\_Alpha\_Hill} over the target modules. We use the default $\gamma_{\mathrm{default}}=\alpha_{\min}(\alpha_{\max}-\alpha_{\min})/v_\alpha$, derived in Appendix~\ref{app:gamma_ablation}. We therefore define the scaled quantization noise of layer $l$ at bit-width $b$ as
		\begin{equation}
			\eta_{l,b} = \left( \frac{\tilde{\alpha}}{{\alpha}_l} \right)^{\gamma}\cdot\mathrm{Var}(\mathbf{W}_l)2^{-2b} .
			\label{eq:sensitivity_metric}
		\end{equation}
	}
	
	{\newupdate
		\noindent\textbf{Budget-Constrained Formulation.}
		Using $\eta_{l,b}$ as the layer-wise cost, we minimize the total scaled quantization noise across all layers under a target bit budget. Let $\mathcal{B}$ be the set of candidate bit-widths (e.g., $\mathcal{B} = \{1, 2, 3, 4\}$). To formalize the allocation decision, we introduce a binary indicator variable $x_{l,b} \in \{0, 1\}$, where $x_{l,b} = 1$ if layer $l$ is assigned bit-width $b$, and $0$ otherwise.
		To ensure a valid configuration, we impose two constraints: i) each layer must be assigned exactly one bit; and ii) for the layer $l$ with $N_l$ parameters, the cost of assigning bit $b$ is $N_l \cdot b$. The total size of the quantized model must not exceed a target budget $B_{\text{tot}}$. 
		
		Combining the objective and constraints yields the following Integer Linear Programming (ILP) formulation:
		\begin{equation}
			\begin{aligned}
				\min_{\{x_{l,b}\}} \quad & \sum_{l=1}^{L} \sum_{b \in \mathcal{B}} x_{l,b} \cdot \eta_{l,b} \\
				\textrm{s.t.} \quad & \sum_{l=1}^{L} \sum_{b \in \mathcal{B}} x_{l,b} \cdot N_l \cdot b \le B_{\text{tot}}, \\
				& \sum_{b \in \mathcal{B}} x_{l,b} = 1, \quad \forall l \in \{1, \dots, L\}, 
				& x_{l,b} \in \{0, 1\}.
			\end{aligned}
			\label{eq:ilp_formulation}
		\end{equation}
		Solving Eq.~\ref{eq:ilp_formulation} provides an optimal bit assignment that minimizes the total importance-scaled quantization noise, and thus the accuracy degradation. Eq.~\ref{eq:ilp_formulation} is a standard instance of the Multiple-Choice Knapsack Problem~\cite{pisinger1998knapsack}. Modern ILP solvers (e.g., PuLP~\cite{mitchell2011pulp}) can solve it to global optimality within seconds.
		
	}

	\section{Empirical Results}
	\subsection{Experimental Setting}
	We evaluate AlphaQ on four representative MoE LLMs: DeepSeekV2-Lite~\cite{liu2024deepseek}, Qwen1.5-MoE~\cite{qwen_moe}, Mixtral-8$\times$7B~\cite{jiang2024mixtral}, and Qwen3-30B-A3B~\cite{yang2025qwen3}.  
	For \texttt{PL\_Alpha\_Hill} calculation, we use FARMS submatrices of size $256 \times 256$ with sampling stride 256 along each dimension.
	For quantization, we follow prior work~\cite{huang2024mixture} and perform weight-only quantization with group-wise, asymmetric GPTQ~\cite{frantar2022gptq} (group size 128, calibrated with 128 samples from WikiText2). We use uniform 4-bit quantization for all non-expert layers, including attention and the router. For experts, we apply mixed-precision quantization at layer granularity: each layer is assigned a bit-width from $\{1,2,3,4\}$, and we control the compression level using the average bits per layer, defined as the average bit-width across all layers in MoE blocks. We report results under two bits-per-layer settings: 2.5 and 3.5. 
	
	\looseness=-1 We compare AlphaQ against two baselines under matched budget settings: Uniform, which assigns the same bit to all experts, and PMQ, a calibration-based mixed-precision MoE quantization method~\cite{huang2024mixture}. 
	Specifically, for Uniform, when the budget is $x.5$, we follow prior work~\cite{huang2024mixture} by setting the first half of the MoE blocks to $x{+}1$ bits and the second half to $x$ bits. Moreover, we compare AlphaQ with additional bit-allocation methods in certain settings, including block score predictor (BSP)~\cite{li2024examining}, a Hessian-based method (Hessian)~\cite{dong2020hawq}, and Automated Fine-Grained MoE Quantization (AFG)~\cite{xie2025automated}.
	For evaluation, we report perplexity (PPL $\downarrow$) on WikiText2 and average zero-shot accuracy (Avg. $\uparrow$) over five benchmarks: PIQA~\cite{piqa}, ARC-Easy (ARC-e), ARC-Challenge (ARC-c)~\cite{arcEasyAndChallenge}, HellaSwag~\cite{hellaswag}, and WinoGrande (WinoG.)~\cite{winogrande}, using the EleutherAI LM Harness~\cite{eval-harness}. 
	
	\subsection{Main Results}
    We find that mixed-precision quantization generally yields higher quality than uniform bit-width assignment, especially under low-bit budgets, reflecting the heterogeneous importance of layers in MoE models. 
    This is shown in Table~\ref{tab:table-moe-main}.
	Moreover, compared with the calibration-based method, AlphaQ achieves consistently stronger results across models and budgets, indicating that our calibration-free bit-allocation method yields better performance. More detailed results are provided in Appendix~\ref{app:further_main}.

\begin{table*}[t]
		\caption{\textbf{Results on DeepSeekV2-Lite, Qwen1.5-MoE, Mixtral-8$\times$7B and Qwen3-30B-A3B.}
			Perplexity$\downarrow$ on WikiText2 and accuracy$\uparrow$ on five zero-shot tasks.
			Avg. denotes the average accuracy across the five tasks.
			The best results under each bit budget are highlighted in bold.}
		\label{tab:table-moe-main}
		\centering
		\small
		\setlength{\tabcolsep}{4pt}
		\renewcommand{\arraystretch}{1.08}
		\resizebox{0.85\linewidth}{!}{
			
			\begin{tabular}{l c l c c c c c c c}
				\toprule
				\multirow{2}{*}{\textbf{Model}} &
				\multirow{2}{*}{\textbf{Bits}} &
				\multirow{2}{*}{\textbf{Method}} &
				\multirow{2}{*}{\textbf{WikiText2$\downarrow$}} &
				\multicolumn{6}{c}{\textbf{Accuracy}$\uparrow$} \\
				\cmidrule(lr){5-10}
				& & &
				\multicolumn{1}{c}{} &
				\textbf{PIQA} & \textbf{ARC-e} & \textbf{ARC-c} &
				\textbf{HellaSwag} & \textbf{WinoG.} & \textbf{Avg.} \\
				\midrule
				\multirow{10}{*}{Mixtral-8$\times$7B}
				& 16 & Original & 3.84 & 83.57 & 83.67 & 59.30 & 84.03 & 76.24 & 77.36 \\
				\cmidrule(lr){2-10}
				\addlinespace[2pt]
				& \multirow{6}{*}{2.5}
				& Uniform  & 6.16  & 79.38 & 77.40 & 50.26 & 58.84 & 73.40 & 67.86 \\
				& & BSP     & 13.61 & 68.23 & 54.97 & 28.38 & 55.61 & 62.19 & 53.88 \\
				& & Hessian & 5.41  & 80.21 & 76.38 & 51.20 & 67.05 & 72.97 & 69.56 \\
				& & AFG     & --    & 74.32 & 69.20 & 37.57 & 67.88 & 63.80 & 62.55 \\
				& & PMQ     & 5.32  & 80.36 & 77.10 & 51.28 & 69.23 & \textbf{73.95} & 70.38 \\
				& & AlphaQ  & \textbf{5.17} & \textbf{80.84} & \textbf{78.87} & \textbf{51.59} & \textbf{69.24} & 73.24 & \textbf{70.76} \\
				\cmidrule(lr){2-10}
				\addlinespace[2pt]
				& \multirow{3}{*}{3.5}
				& Uniform & 4.29 & 81.28 & 83.04 & 54.95 & 73.58 & \textbf{76.19} & 73.81 \\
				& & PMQ     & 4.25 & 81.21 & 82.49 & 53.92 & 72.90 & 76.09 & 73.32 \\
				& & AlphaQ  & \textbf{4.23} & \textbf{81.51} & \textbf{83.38} & \textbf{55.80} & \textbf{74.67} & 76.10 & \textbf{74.29} \\
				\midrule
				\multirow{8}{*}{DeepSeekV2-Lite}
				& 16 & Original & 6.31 & 80.08 & 76.47 & 48.98 & 78.01 & 71.51 & 71.01 \\
				\cmidrule(lr){2-10}
				\addlinespace[2pt]
				& \multirow{4}{*}{2.5}
				& Uniform & 8.14 & 76.66 & 71.34 & 35.84 & 72.44 & 69.46 & 65.15 \\
				& & AFG     & --   & 74.32 & 69.20 & 37.57 & 67.88 & 63.80 & 62.55 \\
				& & PMQ     & 6.95 & 77.86 & 71.68 & 37.37 & 72.76 & 69.06 & 65.75 \\
				& & AlphaQ  & \textbf{6.64} & \textbf{78.02} & \textbf{73.77} & \textbf{39.22} & \textbf{73.38} & \textbf{69.30} & \textbf{66.74} \\
				\cmidrule(lr){2-10}
				\addlinespace[2pt]
				& \multirow{3}{*}{3.5}
				& Uniform & 6.83 & 79.21 & 74.56 & 44.57 & 76.27 & 70.83 & 69.09 \\
				& & PMQ     & 6.57 & 79.63 & 75.26 & 45.83 & 76.97 & 71.04 & 69.75 \\
				& & AlphaQ  & \textbf{6.44} & \textbf{79.89} & \textbf{75.51} & \textbf{46.51} & \textbf{77.21} & \textbf{71.21} & \textbf{70.07} \\
				\midrule
				
				\multirow{8}{*}{Qwen1.5-MoE}
				& 16 & Original & 7.22 & 80.52 & 69.23 & 44.11 & 77.21 & 68.59 & 67.93 \\
				\cmidrule(lr){2-10}
				\addlinespace[2pt]
				& \multirow{4}{*}{2.5}
				& Uniform & 10.36 & 71.49 & 49.07 & 29.44 & 53.98 & 55.96 & 51.99 \\
				& & AFG     & --    & 77.53 & 61.07 & 31.20 & \textbf{72.77} & 62.12 & 60.94 \\
				& & PMQ     & 9.01  & 77.58 & 60.56 & 36.26 & 70.69 & 65.82 & 62.18 \\
				& & AlphaQ  & \textbf{8.14} & \textbf{78.21} & \textbf{66.97} & \textbf{42.72} & 71.81 & \textbf{66.12} & \textbf{65.17} \\
				\cmidrule(lr){2-10}
				\addlinespace[2pt]
				& \multirow{3}{*}{3.5}
				& Uniform & 8.13 & 79.31 & 66.81 & 41.51 & 74.52 & 67.54 & 65.94 \\
				& & PMQ     & 7.42 & 80.31 & 69.02 & 43.81 & 77.13 & 68.50 & 67.75 \\
				& & AlphaQ  & \textbf{7.38} & \textbf{80.46} & \textbf{69.15} & \textbf{44.33} & \textbf{77.70} & \textbf{68.55} & \textbf{68.04} \\
				\midrule

				\multirow{5}{*}{Qwen3-30B-A3B}
				& 16 & Original & 8.71 & 80.25 & 79.29 & 55.80 & 77.75 & 71.11 & 72.84 \\
				\cmidrule(lr){2-10}
				\addlinespace[2pt]
				& \multirow{3}{*}{2.5}
				& Uniform & 10.29 & 78.07 & \textbf{75.93} & 50.07 & 71.72 & 68.01 & 68.76 \\
				& & PMQ     & 10.94 & 76.61 & 68.43 & 45.99 & 72.13 & 66.22 & 65.88 \\
				& & AlphaQ    & \textbf{9.27} & \textbf{78.34} & 75.87 & \textbf{50.35} & \textbf{74.49} & \textbf{68.71} &  \textbf{69.56}\\
				\bottomrule
			\end{tabular}
		}
	\end{table*}
    
	{\newupdate
		Notably, on Qwen1.5-MoE, AlphaQ with a bit budget of 3.5 performs competitively with the BF16 model in average accuracy over five zero-shot tasks. 
		In the more aggressive 2.5-bit setting, AlphaQ also maintains relatively strong performance across all three models.
		For example, its average accuracy on zero-shot tasks drops by only 4.3\% on DeepSeekV2-Lite, 2.8\% on Qwen1.5-MoE, and 6.6\% on Mixtral-8$\times$7B, demonstrating that our bit allocation remains beneficial even when the bit budget is extremely tight.
		To evaluate AlphaQ on larger-scale MoE models, we include results for Qwen3-30B-A3B. Under a 2.5-bit budget, Qwen3-30B-A3B also surpasses both uniform and PMQ in overall performance.
		
	}
	
	{\newupdate
		\subsection{AlphaQ vs. Multi-Domain Calibration Baselines}
		To further compare AlphaQ to baselines with broader calibration coverage, we use PMQ~\cite{huang2024mixture} on OLMoE-1B-7B~\cite{muennighoff2024olmoe} under a 3-bit budget with various single-domain and mixed-domain calibration sets, and we evaluate the quantized models on the common-sense benchmark MMLU, the math reasoning benchmark MATH, and the Chinese language benchmark CEval~\cite{huang2023ceval}.
	}
	
	{\newupdate
		\begin{table*}[t]
			\centering
			\begin{minipage}[t]{0.49\textwidth}
				\centering
				\caption{\textbf{Comparison between AlphaQ and PMQ with multi-domain calibration on OLMoE-1B-7B under a 3-bit budget.} The best results in each benchmark are highlighted in bold.}
				\label{tab:olmoe_multidomain_pmq}
				\small
				\setlength{\tabcolsep}{5pt}
				\renewcommand{\arraystretch}{1.15}
				\resizebox{\linewidth}{!}{
					\begin{tabular}{l c c c c}
						\toprule
						\textbf{Method} & \textbf{MMLU} & \textbf{MATH} & \textbf{CEval} & \textbf{Avg.} \\
						\midrule
						BF16 & 53.40 & 21.40 & 31.06 & 35.29 \\
						\midrule
						Uniform 3-bit & 43.56 & 11.45 & 27.19 & 27.40 \\
						PMQ (C4) & 43.80 & 12.07 & 25.11 & 26.99 \\
						PMQ (CEval-dev) & 44.04 & 11.15 & 22.66 & 25.95 \\
						PMQ (MATH) & 44.43 & 12.82 & 22.96 & 26.74 \\
						PMQ (C4+CEval-dev) & \textbf{44.79} & 11.71 & 23.03 & 26.51 \\
						PMQ (C4+MATH) & 44.38 & 12.91 & 25.78 & 27.69 \\
						PMQ (MATH+CEval-dev) & 44.33 & 12.37 & 21.99 & 26.23 \\
						AlphaQ & 44.75 & \textbf{13.11} & \textbf{28.72} & \textbf{28.86} \\
						\bottomrule
					\end{tabular}
				}
			\end{minipage}
			\hfill
			\begin{minipage}[t]{0.49\textwidth}
				\centering
				\caption{\textbf{Component ablation on OLMoE-1B-7B.} We report perplexity on WikiText2 and average accuracy over six zero-shot tasks. Alpha denotes \texttt{PL\_Alpha\_Hill}, and \textbf{Direct} means multiplying Alpha and Quantization Noise directly.}
				\label{tab:component_ablation}
				\scriptsize
				\setlength{\tabcolsep}{3pt}
				\renewcommand{\arraystretch}{1.03}
				\resizebox{0.9\linewidth}{!}{
					\begin{tabular}{lcc}
						\toprule
						\textbf{Method} & \textbf{PPL $\downarrow$} & \textbf{Avg. Acc. $\uparrow$} \\
						\midrule
						BF16 & 7.95 & 70.45 \\
						Uniform 3-bit & 11.78 & 64.73 \\
						Noise-only & 11.22 & 63.74 \\
						Alpha-only & 10.03 & 66.23 \\
						Alpha+Noise (Direct) & 9.56 & 66.81 \\
						Alpha+Noise (Ours) & \textbf{9.19} & \textbf{67.11} \\
						\bottomrule
					\end{tabular}
				}
			\end{minipage}
		\end{table*}
		
		Table~\ref{tab:olmoe_multidomain_pmq} shows that PMQ remains sensitive to the calibration domains even when multi-domain mixtures are allowed.
		Among the calibration setting variants, even the multi-domain PMQ configuration does not consistently match Uniform on every axis, whereas AlphaQ improves the overall average by avoiding calibration-domain bias. 
		We also note that AlphaQ performs better than the recent cross-dataset method DynaMo on OLMoE benchmarks.
        Detailed results are deferred to Appendix~\ref{app:dynamo_compare}.
		
		These results show that enlarging calibration coverage shifts the failure mode across tasks, rather than removing sensitivity to the calibration datasets. 
		
		\subsection{Ablation Study}
		
		\paragraph{Which Components Matter in AlphaQ?}
		To isolate the contribution of each component, we perform an ablation study on OLMoE-1B-7B under a 3-bit budget. Our goal is to quantify the gains from \texttt{PL\_Alpha\_Hill}, quantization noise, and their combination.
	}

	{\newupdate
		\looseness=-1 Table~\ref{tab:component_ablation} shows that \texttt{PL\_Alpha\_Hill} already provides a much stronger signal than quantization noise for bit allocation, indicating that \texttt{PL\_Alpha\_Hill} captures meaningful layer importance.
		However, neither \texttt{PL\_Alpha\_Hill} nor noise-only allocation is sufficient, as both remain clearly behind the final AlphaQ solution. 
	}
	
	{\newupdate
		Incorporating quantization noise improves performance further. The gap between the Direct variant and the final variant shows that the way \texttt{PL\_Alpha\_Hill} and quantization noise are combined also matters. The best results are obtained only when the alpha-based importance metric and the quantization-noise term are properly used together.
	}
	
	\paragraph{How to Allocate Bit-Width Across Blocks?}
	\begin{table}[t]
		\centering
		\begin{minipage}[t]{0.49\linewidth}
			\centering
			\caption{\textbf{Ablation study for bit-budget allocation.} We compare perplexity (PPL) on WikiText2 and accuracy (Acc.) on six zero-shot benchmarks under block-wise and global budgets.}
			\label{tab:budget-allocation}
			\small
			\setlength{\tabcolsep}{8pt}
			\renewcommand{\arraystretch}{1.2}
			\resizebox{\linewidth}{!}{
				\begin{tabular}{c l | cc cc}
					\toprule
					\multirow{2}{*}{Bit} & \multirow{2}{*}{Budget}
					& \multicolumn{2}{c}{Mixtral-8$\times$7B}
					& \multicolumn{2}{c}{DeepSeekV2-Lite} \\
					\cmidrule(lr){3-4}\cmidrule(lr){5-6}
					& & PPL ($\downarrow$) & Acc. ($\uparrow$) & PPL ($\downarrow$) & Acc. ($\uparrow$) \\
					
					\midrule
					\multirow{2}{*}{2.5} & Block  & 5.81  & 71.68 & 6.99 & 64.61 \\
					& Global (Ours) & \textbf{5.17}  & \textbf{72.39}& \textbf{6.64} & \textbf{65.69} \\
					\midrule
					\multirow{2}{*}{3.5}   & Block  & 4.71  & 72.56 & 6.59 & 68.72 \\
					& Global (Ours) & \textbf{4.23} & \textbf{75.98} & \textbf{6.44} & \textbf{69.03} \\
					\bottomrule
				\end{tabular}
			}
		\end{minipage}
		\hfill
		\begin{minipage}[t]{0.49\linewidth}
			\centering
			\caption{\textbf{Ablation study for bit-allocation granularity.} We compare perplexity on WikiText2 in expert-wise and layer-wise bit allocation under different bit budgets.}
			\label{tab:bit-allocation-granularity}
			\small
			\setlength{\tabcolsep}{8pt}
			\renewcommand{\arraystretch}{1.2}
			\resizebox{\linewidth}{!}{
				\begin{tabular}{l | cc | cc}
					\toprule
					\textbf{Model} 
					& \multicolumn{2}{c|}{\textbf{2-bit PPL $\downarrow$}} 
					& \multicolumn{2}{c}{\textbf{3-bit PPL $\downarrow$}} \\
					\cmidrule(lr){2-3} \cmidrule(lr){4-5}
					& Expert & Layer & Expert & Layer \\
					\midrule
					Mixtral 8$\times$7B 
					& 6.28 & \textbf{6.11} 
					& 4.72 & \textbf{4.37} \\
					DeepSeekV2-Lite 
					& 7.47 & \textbf{7.30} 
					& 6.81 & \textbf{6.69} \\
					\bottomrule
				\end{tabular}
			}
		\end{minipage}
		
	\end{table}

	\looseness=-1 We conduct an ablation study on two budget-allocation strategies: i) fixing the global average bit-width for the entire model; and ii) fixing the average bit-width for each block. The results are reported in Table~\ref{tab:budget-allocation}. In all settings, global budgeting outperforms block-wise budgeting. This indicates that different blocks in MoE models have unequal importance, and HT-SR theory provides a unified view for comparing layer importance across blocks.

	\paragraph{How to Allocate Bit-Width within Blocks?}

	We perform an ablation study on bit-allocation granularity, comparing layer-wise and expert-wise allocation. As shown in Table~\ref{tab:bit-allocation-granularity}, the finer-grained, layer-wise strategy consistently outperforms expert-wise allocation across all settings. These findings indicate that AlphaQ can effectively capture the relative importance of layers within each expert and allocate bit-widths accordingly, enabling finer-grained mixed-precision quantization.
	\subsection{Efficiency Evaluation}

	{\newupdate
		\paragraph{End-to-end efficiency.}
		We evaluate AlphaQ's efficiency on an NVIDIA A40 GPU (48\,GB). Figure~\ref{fig:efficiency_speed_memory} reports the trade-offs among inference speedup, memory footprint, and accuracy.
		\begin{figure}[t]
			\centering
			\begin{subfigure}[t]{0.47\linewidth}
				\centering
				\includegraphics[width=0.95\linewidth]{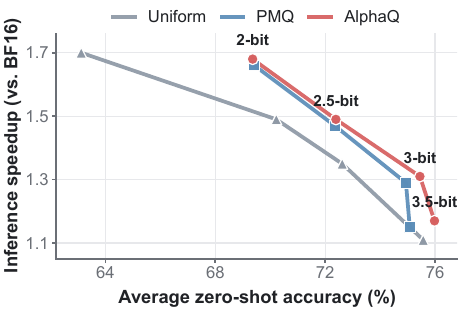}
				\caption{Accuracy--speedup on Mixtral-8$\times$7B.}
				\label{fig:pareto}
			\end{subfigure}
			\hfill
			\begin{subfigure}[t]{0.48\linewidth}
				\centering
				\includegraphics[width=1\linewidth]{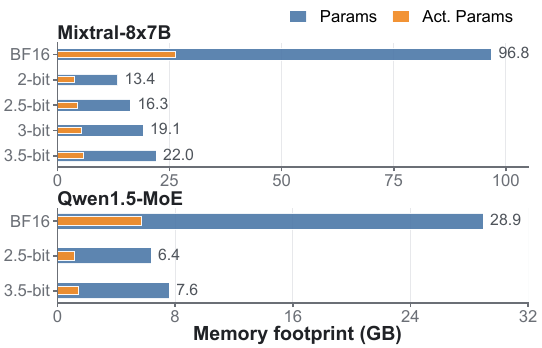}
				\caption{Memory footprint of AlphaQ.}
				\label{fig:memory_footprint}
			\end{subfigure}
			\caption{\textbf{End-to-end efficiency of AlphaQ.} Left: average zero-shot accuracy versus inference speedup relative to BF16 for varying bit budgets on Mixtral-8$\times$7B. Right: parameter memory footprint of Mixtral-8$\times$7B and Qwen1.5-MoE. }
			\label{fig:efficiency_speed_memory}
		\end{figure}
		AlphaQ consistently dominates the frontier for Mixtral-8$\times$7B, providing a superior trade-off between accuracy and inference speedup. Furthermore, AlphaQ substantially alleviates the MoE memory bottleneck across bit budgets. For Mixtral-8$\times$7B, AlphaQ reaches 1.68$\times$ speedup at 2-bit while reducing parameter memory from 96.8\,GB to 13.4\,GB. Even at a 3.5-bit budget, it reduces Mixtral-8$\times$7B parameter memory to 22.0\,GB, corresponding to 4.4$\times$ compression. For Qwen1.5-MoE at a 3.5-bit budget, AlphaQ preserves BF16-level accuracy with only 7.6\,GB of weights, about one quarter of BF16 memory. More detailed efficiency results are provided in Appendix~\ref{app:acc-speed}.
	}
	
	{\newupdate
		\paragraph{PL\_Alpha\_Hill overhead.}
		Computing \texttt{PL\_Alpha\_Hill} is an offline preprocessing step, performed once per model before solving the ILP.
		As shown in Table~\ref{tab:alpha_time}, this cost is modest: one full alpha-computation pass takes about three minutes for OLMoE-1B-7B and about nine minutes for Qwen3-30B-A3B, with only 44--57 ms per module.
		Since this step is performed before quantization and does not affect online inference, we consider the overhead practical for deployment.
		\begin{table}[t]
			\centering
			\caption{\textbf{Offline computation cost of \texttt{PL\_Alpha\_Hill}.} We report the total time and the average per-module time for one full alpha-computation pass.}
			\label{tab:alpha_time}
			\resizebox{0.78\linewidth}{!}{
				\begin{tabular}{l c c c}
					\toprule
					\textbf{Model} & \textbf{\# Modules} & \textbf{Total Time (s)} & \textbf{Per-Module (ms)} \\
					\midrule
					OLMoE-1B-7B & 3152 & 180.28 & 57.19 \\
					Qwen3-30B-A3B & 12448 & 549.26 & 44.12 \\
					\bottomrule
			\end{tabular}}
		\end{table}
	}
	
	\paragraph{Quantized inference backend.}
	We implement a quantized inference backend that combines low-bit dequantization with Tensor Core GEMM.
	Relative to the original BF16 model, this backend improves both prefill and decode.
	We reuse the PMQ-style HQQ~\cite{badri2023hqq} backend for dequantization in the prefill stage of all layers and the decode stage of non-expert layers, and implement fused Triton kernels to reduce memory traffic in the decode stage of MoE expert layers.
	Detailed runtime implementation and profiling are provided in Appendix~\ref{app:decode_runtime}.

	\section{Conclusion}
	\looseness=-1 In this work, we propose \textbf{AlphaQ}, a calibration-free bit-allocation method for MoE quantization. Motivated by HT-SR theory, AlphaQ derives an importance signal directly from model weights by measuring spectral heavy-tailedness. Using this metric, we systematically analyze MoE models from an HT-SR perspective and reveal importance diversity across multiple levels. Based on these observations, AlphaQ allocates bit-widths via a global constrained optimization under a fixed bit budget, enabling calibration-free bit allocation for MoE models. Extensive experiments show that weight-based bit allocation is an effective and promising approach for improving the generalization of mixed-precision MoE quantization.
	
	
	\bibliography{alphaq}
	\bibliographystyle{plainnat}
	
	\newpage
	\FloatBarrier
	\appendix
	\section{Appendix}
	\subsection{Calibration-dependent Expert Activation and Bit-Width Allocation}
	\label{app:activation_pattern}
\begin{figure*}
		\centering
		\includegraphics[width=1\linewidth]{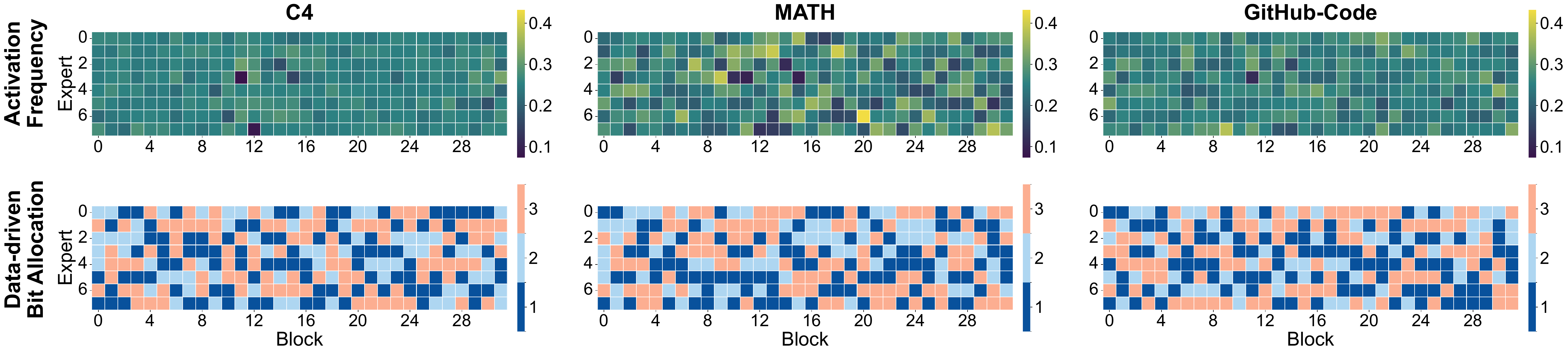}
		\caption{\textbf{Domain-dependent expert activation patterns and data-driven bit-width allocation in Mixtral-8$\times$7B.} Activation frequencies (top) and corresponding bit-width allocations (bottom) across different domains (C4, MATH, GitHub-Code), illustrating substantial variations induced by calibration data from different domains.}
		\label{fig:heatmap}
	\end{figure*}

	To illustrate the influence of calibration datasets on MoE bit-width allocation, we apply PMQ~\cite{huang2024mixture} to Mixtral-8$\times$7B using datasets from different domains. As illustrated in Figure~\ref{fig:heatmap}, the collected expert activation frequencies vary significantly as a function of the calibration data. As a result, different calibration-time statistics lead to substantially different data-driven allocations.
	\subsection{Terminology Summary}
	\label{app:teminology_summary}
	
\begin{figure}
		\centering
		\includegraphics[width=0.6\linewidth]{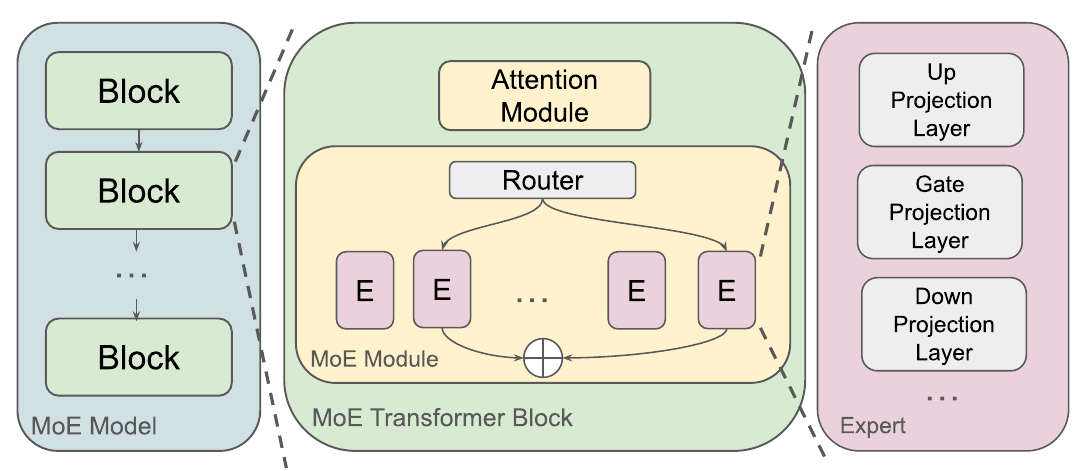}
		\caption{\textbf{Hierarchical relationship among blocks, experts, and layers in our paper.} An MoE model consists of multiple Transformer blocks; each block contains an attention module and an MoE module with multiple experts; each expert further comprises multiple layers (e.g., up, gate, and down projection).}
		\label{fig:terminology}
	\end{figure}

	To avoid ambiguity, we summarize the terminology used throughout this paper in \cref{fig:terminology}; the descriptions are as follows:
	\begin{itemize}[leftmargin=*]
		\item \textbf{Global:} Global refers to the entire model. A global budget means that a single bit budget is defined for the whole model and shared across all blocks, experts, and layers.
		
		\item \textbf{Block:} A block refers to a Transformer block. In popular MoE models, each block consists of three main modules: an attention module, a MoE module containing multiple experts, and a routing module. A block-wise budget assigns the same bit budget independently to each Transformer block.
		
		\item \textbf{Expert:} An expert is a feed-forward network within the MoE module. Each expert typically contains multiple layers (e.g., up/gate/down projections). An expert-wise bit allocation means that all submodules within an expert share a single bit setting.
		
		\item \textbf{Layer:} A layer denotes an individual submodule within a module, such as projection layers (e.g., up, gate, and down projections in experts). A layer-wise bit allocation means assigning independent bit-widths to each submodule.

	\end{itemize}
	
	\subsection{Derivation of the Hill Estimator}
	\label{app:hill-derivation}
	
	\subsubsection{Tail Model Assumption}
	Following Section~\ref{sec:Alpha_Score}, we model the ESD tail by a truncated power-law \emph{density}:
	\begin{equation}
		p(\lambda) \propto \lambda^{-\alpha}, \qquad \lambda_{\min} < \lambda < \lambda_{\max},
		\label{eq:pl_density}
	\end{equation}
	where $\alpha$ is the power-law exponent. Smaller $\alpha$ indicates heavier tails.
	
	\subsubsection{From the Power-Law Density to a Pareto Form}
	To derive the estimator used in Eq.~\ref{eq:pl_alpha_hill}, we rewrite Eq.~\ref{eq:pl_density} as a Pareto distribution.
	Let $\beta = \alpha - 1$. Then the density can be expressed as
	\begin{equation}
		p(\lambda)
		=
		\beta\,\lambda_{\min}^{\beta}\,\lambda^{-(\beta+1)},
		\qquad \lambda \ge \lambda_{\min},
		\label{eq:pareto_pdf}
	\end{equation}
	where the normalization constant is determined by
	\begin{equation}
		1
		=
		\int_{\lambda_{\min}}^{\infty} \beta\,\lambda_{\min}^{\beta}\,\lambda^{-(\beta+1)}\, d\lambda.
	\end{equation}
	Eq.~\ref{eq:pareto_pdf} is the standard Pareto form, and estimating $\beta$ yields $\alpha=\beta+1$.
	
	\subsubsection{Likelihood and Log-Likelihood}
	Let $\lambda_1,\dots,\lambda_k$ be $k$ independent tail samples with $\lambda_i \ge \lambda_{\min}$.
	Under Eq.~\ref{eq:pareto_pdf}, the likelihood of $\beta$ is
	\begin{equation}
		L(\beta)
		=
		\prod_{i=1}^{k} \beta\,\lambda_{\min}^{\beta}\,\lambda_i^{-(\beta+1)}
		=
		\beta^{k}\lambda_{\min}^{k\beta}\prod_{i=1}^{k}\lambda_i^{-(\beta+1)}.
	\end{equation}
	Taking logs gives
	\begin{equation}
		\ln L(\beta)
		=
		k\ln\beta
		+
		k\beta\ln\lambda_{\min}
		-
		(\beta+1)\sum_{i=1}^{k}\ln\lambda_i.
	\end{equation}
	
\begin{figure}
		\centering
		\includegraphics[width=1\linewidth]{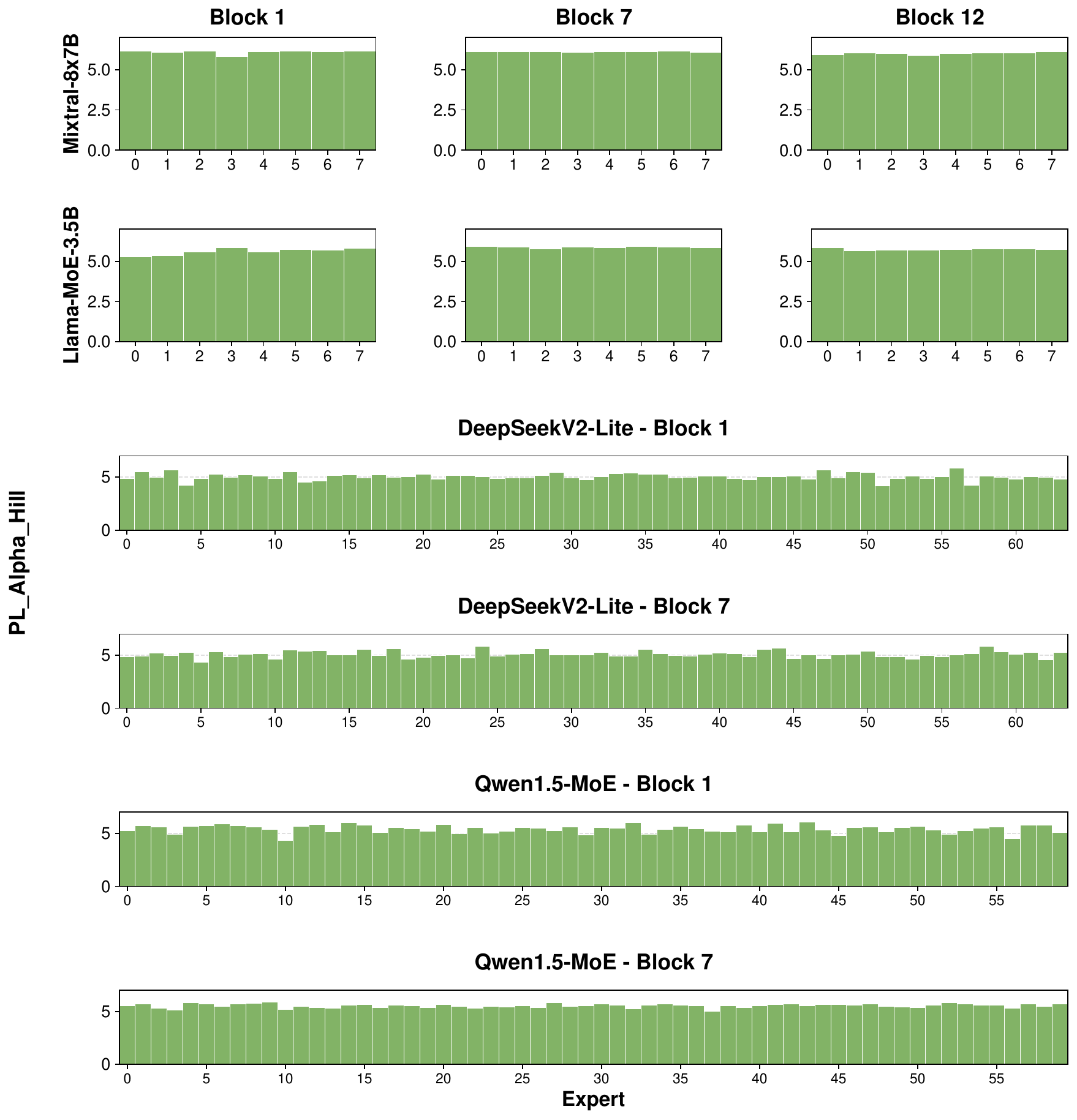}
		\caption{\textbf{Expert-wise \texttt{PL\_Alpha\_Hill} distribution in sampled MoE blocks.} Experts within the same MoE layer exhibit different alpha values, indicating that expert importance is heterogeneous even within a single block.}
		\label{fig:block_alpha_barchart_expert}
	\end{figure}

\begin{figure}
		\centering
		\includegraphics[width=1\linewidth]{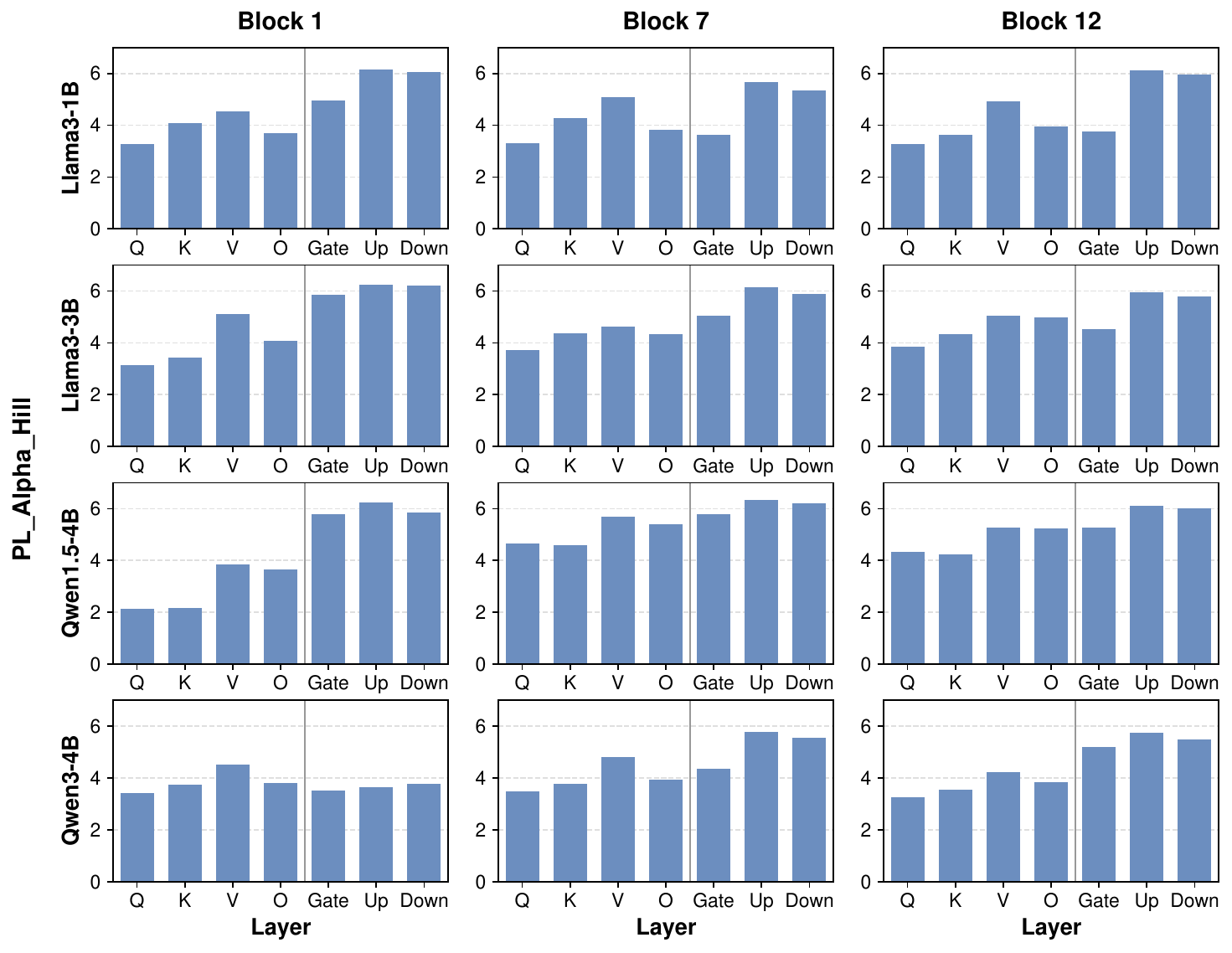}
		\caption{Layer-wise \texttt{PL\_Alpha\_Hill} distribution of sampled blocks in four non-MoE LLMs}
		\label{fig:dense_alpha_barchart}
	\end{figure}

	{\newupdate
		\subsubsection{Maximum Likelihood Estimation}
		Differentiating and setting the derivative to zero,
		\begin{equation}
			\frac{\partial \ln L}{\partial \beta}
			=
			\frac{k}{\beta}
			+
			k\ln\lambda_{\min}
			-
			\sum_{i=1}^{k}\ln\lambda_i
			=0,
		\end{equation}
		which yields
		\begin{equation}
			\frac{1}{\beta}
			=
			\frac{1}{k}\sum_{i=1}^{k}\ln\!\left(\frac{\lambda_i}{\lambda_{\min}}\right).
			\label{eq:beta_mle}
		\end{equation}
		Therefore,
		\begin{equation}
			\hat{\beta}
			=
			\left(
			\frac{1}{k}\sum_{i=1}^{k}\ln\!\left(\frac{\lambda_i}{\lambda_{\min}}\right)
			\right)^{-1},
			\qquad
			\hat{\alpha}
			=
			1+\hat{\beta}.
			\label{eq:alpha_from_beta}
		\end{equation}
		In our implementation, we concatenate eigenvalues across submatrices, and we sort them in ascending order:
		\begin{equation}
			\lambda_1 \le \lambda_2 \le \cdots \le \lambda_n.
		\end{equation}
		We take the top-$k$ eigenvalues $\{\lambda_{n-k+1},\dots,\lambda_n\}$ as tail samples and set the
		lower cutoff as the $(n-k)$-th order statistic, i.e., $\lambda_{\min}=\lambda_{n-k}$.
		Substituting $\lambda_i \leftarrow \lambda_{n-i+1}$ and $\lambda_{\min}=\lambda_{n-k}$ into
		Eq.~\ref{eq:alpha_from_beta} gives the Hill estimator used in Eq.~\ref{eq:pl_alpha_hill}:
		\begin{equation}
			\hat{\alpha}
			=
			1+
			\left(
			\frac{1}{k}\sum_{i=1}^{k}\ln\frac{\lambda_{n-i+1}}{\lambda_{n-k}}
			\right)^{-1}.
			\label{eq:hill_final}
		\end{equation}
		\noindent\textbf{Intuition.}
		The estimator is the reciprocal of the average log-excess above the threshold. Heavier tails produce larger log-excess values and smaller $\hat{\alpha}$. Lighter tails produce smaller log-excess values and larger $\hat{\alpha}$.
		
		\noindent\textbf{FARMS-based score computation.}
		To make the spectral estimate comparable across heterogeneous matrix shapes, we compute \texttt{PL\_Alpha\_Hill} with the \emph{Fixed-Aspect-Ratio Matrix Subsampling} (FARMS) heuristic~\cite{hu2025eigenspectrum}. For each layer, we partition its weight matrix into submatrices with a fixed aspect ratio, form the corresponding Gram matrices, and collect their eigenvalues. We then concatenate these eigenvalues, sort them in ascending order, and apply the Hill estimator in Eq.~\ref{eq:hill_final}. This yields a more shape-robust estimate of spectral heavy-tailedness.
		
	}
	{\newupdate
		\subsection{Additional \texttt{PL\_Alpha\_Hill} Comparisons}
		\label{app:more_detailed_alpha}
		Section~\ref{sec:pl_alpha} reports layer-wise \texttt{PL\_Alpha\_Hill} distributions that motivate AlphaQ for MoE bit allocation.
		We report further expert-wise \texttt{PL\_Alpha\_Hill} distributions within sampled blocks in Figure~\ref{fig:block_alpha_barchart_expert}.
		These results show that experts within the same MoE layer can exhibit different \texttt{PL\_Alpha\_Hill} values, indicating that expert importance is heterogeneous even within a single block.
		For comparison, Figure~\ref{fig:dense_alpha_barchart} reports layer-wise \texttt{PL\_Alpha\_Hill} distributions for four non-MoE models, including Llama3-1B~\cite{grattafiori2024llama}, Llama3-3B, Qwen1.5-4B~\cite{qwen1.5}, and Qwen3-4B~\cite{yang2025qwen3}.
		These dense-model results serve as supplementary evidence that the strong expert- and block-level heterogeneity analyzed in Section~\ref{sec:pl_alpha} is a particularly important issue for MoE quantization.
	}
	
	{\newupdate
		\subsection{Importance and Quantization Noise Analysis}
		\label{app:alpha_noise_analysis}
		To construct the module-level analysis, we sample projection modules and quantize one sampled module at a time to 2-bit while keeping the remaining weights full-precision. We then measure the resulting perplexity increase as the module-level degradation. The quantization-noise axis uses our quantization noise model in~\cref{sec:bit_allocation_opt}. Since smaller \texttt{PL\_Alpha\_Hill} indicates heavier-tailed spectra and higher structural importance, AlphaQ predicts that large degradation should occur when high quantization noise overlaps with low \texttt{PL\_Alpha\_Hill}.
		\begin{figure}[t]
			\centering
			\includegraphics[width=1\linewidth]{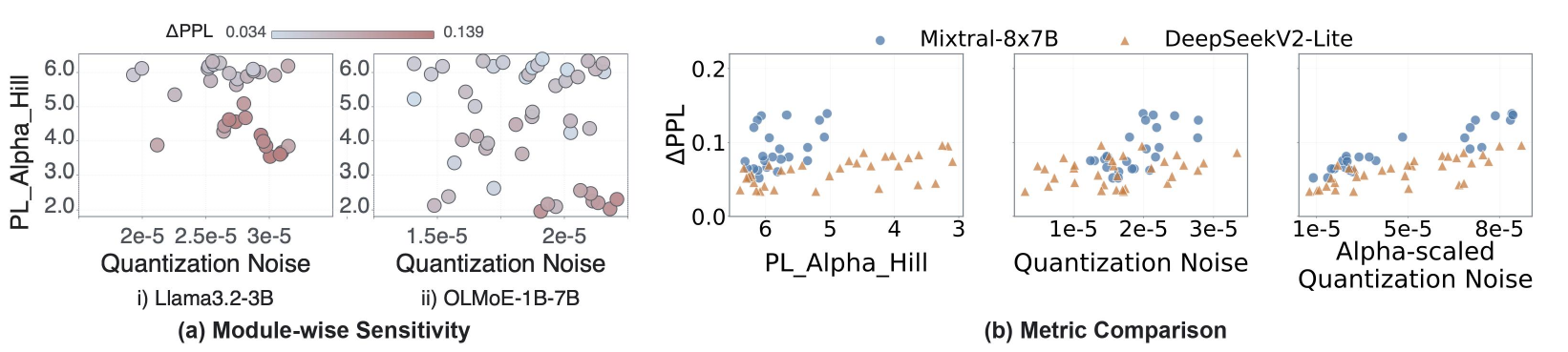}
			\caption{\textbf{Module-level relationship between PL\_Alpha\_Hill, quantization noise, and quantization degradation.} (a) Each point denotes a sampled module from Llama 3.2-3B or OLMoE-1B-7B. The horizontal axis is 2-bit quantization noise, the vertical axis is \texttt{PL\_Alpha\_Hill}, and darker points indicate larger PPL increase after 2-bit quantization. Severe degradation concentrates in the region with high quantization noise and low \texttt{PL\_Alpha\_Hill}. (b) Relationship between PPL degradation and three metrics on sampled modules in Mixtral-8$\times$7B and DeepSeek-V2-Lite: \texttt{PL\_Alpha\_Hill}, Quantization Noise, and Alpha-scaled Quantization Noise. Among the three, Alpha-scaled Quantization Noise shows the clearest monotonic trend with degradation.}
			\label{fig:alpha_noise_joint}
		\end{figure}
		
		Figure~\ref{fig:alpha_noise_joint} shows the joint view of \texttt{PL\_Alpha\_Hill} and quantization noise at the module level. Panel (a) visualizes this interaction directly, while Panel (b) compares individual and combined metrics. Large degradation tends to concentrate in the regime of high quantization noise and low \texttt{PL\_Alpha\_Hill}, but the spread of points also indicates that neither signal alone is sufficient. The clearer monotonic trend of Alpha-scaled Quantization Noise supports the form of our objective: quantization noise should be weighted by structural importance rather than used alone. This is consistent with our component ablation in Table~\ref{tab:component_ablation}.
	}

	\subsection{Justification of the Quantization Noise Model}
	\label{app:quant_noise_model}
	
	Here, we provide a theoretical justification for modeling the layer-wise quantization error variance as $\eta_{l,b} \propto 2^{-2b}$.
	We consider a uniform quantizer applied to the weights of the $l$-th layer, denoted by $\mathbf{W}_l$. We assume the weights lie within the interval $[-R_l, R_l]$, where $R_l$ is a clipping value determined by the distribution of $\mathbf{W}_l$. For $b$-bit quantization, the step size $\Delta_{l,b}$ is defined as
	\begin{equation}
		\Delta_{l,b} = \frac{2R_l}{2^b}.
	\end{equation}
	Under Bennett’s high-resolution quantization hypothesis~\cite{widrow2008quantization}, when the step size is small enough relative to the signal variation, the quantization error $E = Q(w) - w$ can be approximated as a random variable uniformly distributed over $\left[-\frac{\Delta_{l,b}}{2}, \frac{\Delta_{l,b}}{2}\right]$.
	For a continuous uniform distribution with width $\Delta_{l,b}$, the variance is $\Delta_{l,b}^2/12$. Consequently, the variance of the quantization error in layer $l$ can be approximated as
	\begin{equation}
		\mathrm{Var}(\mathbf{E}_{l,b}) \approx \frac{\Delta_{l,b}^2}{12}
		= \frac{1}{12}\left(\frac{2R_l}{2^b}\right)^2
		= \frac{R_l^2}{3} \cdot 2^{-2b}.
	\end{equation}
	We therefore define $c_l = R_l^2/3$. Since the clipping range $R_l$ is typically proportional to the standard deviation of the weights, it follows that $c_l$ scales with $\mathrm{Var}(\mathbf{W}_l)$. This leads to the exponential decay model used in the main text,
	\begin{equation}
		\eta_{l,b} = c_l  2^{-2b},
	\end{equation}
	where $c_l$ captures the layer-specific scale. 
	
	{\newupdate
		\subsection{Data-Free Default and Sensitivity of \texorpdfstring{$\gamma$}{gamma}}
		\label{app:gamma_ablation}
		
		\begin{figure}[t]
			\centering
			\includegraphics[width=0.8\linewidth]{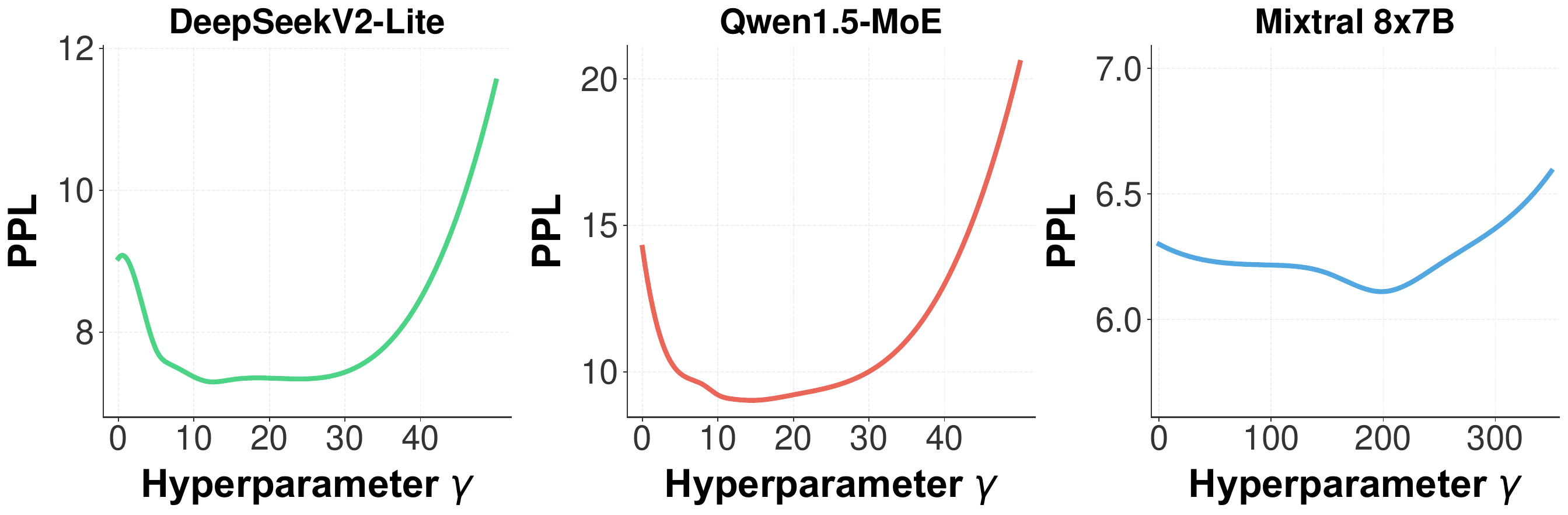}
			\caption{\textbf{Sensitivity analysis of $\gamma$}. Sensitivity varies across MoE models: DeepSeekV2-Lite and Qwen1.5-MoE are more sensitive to $\gamma$ than Mixtral-8$\times$7B.}
			\label{fig:gamma_ablation}
		\end{figure}
		
		We set $\gamma$ from model-weight statistics. Let $\alpha_{\min}$, $\alpha_{\max}$, and $v_\alpha$ denote the minimum, maximum, and variance of \texttt{PL\_Alpha\_Hill} over the target modules. Smaller $\alpha$ indicates stronger heavy-tailedness, so $\alpha_{\min}$ and $\alpha_{\max}$ define model-internal high- and low-importance endpoints.
		
		To set the scale of the power-law mapping $w_\gamma(\alpha)=(\tilde{\alpha}/\alpha)^\gamma$, where $\tilde{\alpha}$ is the mean of $\{\alpha_l\}_l$, we construct a local discriminative proxy between the two endpoints. Specifically, we introduce two local density surrogates
		\begin{equation}
			p_{\text{high}}(\alpha) \propto \exp\left(-\frac{(\alpha - \alpha_{\min})^2}{2v_\alpha}\right), \quad
			p_{\text{low}}(\alpha) \propto \exp\left(-\frac{(\alpha - \alpha_{\max})^2}{2v_\alpha}\right),
		\end{equation}
		which share a common variance scale $v_\alpha$ and are centered at the high- and low-importance endpoints. This construction serves as a local proxy and does not assume a global parametric form of the $\alpha$ distribution.
		
		We define the log-discriminant score
		\begin{equation}
			s(\alpha) = \log \frac{p_{\text{high}}(\alpha)}{p_{\text{low}}(\alpha)}.
		\end{equation}
		Substituting the surrogate forms gives
		\begin{equation}
			s(\alpha)
			=
			-\frac{(\alpha - \alpha_{\min})^2}{2v_\alpha}
			+
			\frac{(\alpha - \alpha_{\max})^2}{2v_\alpha}.
		\end{equation}
		Expanding the squares and simplifying yields
		\begin{equation}
			s(\alpha)
			=
			-\frac{\alpha_{\max} - \alpha_{\min}}{v_\alpha}\cdot \alpha
			+
			C,
		\end{equation}
		where $C$ is a constant independent of $\alpha$. Therefore,
		\begin{equation}
			\frac{d}{d\alpha}s(\alpha)
			=
			-\frac{\alpha_{\max} - \alpha_{\min}}{v_\alpha}.
		\end{equation}
		
		The discriminant score $s(\alpha)$ defines a data-free reference scale describing how rapidly importance changes between the two endpoints. We use this reference to calibrate the curvature of the power-law mapping $w_\gamma(\alpha)$.
		
		The log-slope of the power-law mapping is
		\begin{equation}
			\frac{d}{d\alpha}\log w_\gamma(\alpha)
			=
			-\frac{\gamma}{\alpha}.
		\end{equation}
		
		We match the local slope of $\log w_\gamma(\alpha)$ to the slope of $s(\alpha)$ at the high-importance endpoint $\alpha_{\min}$, where the mapping is most sensitive:
		\begin{equation}
			\left.
			\frac{d}{d\alpha}\log w_\gamma(\alpha)
			\right|_{\alpha=\alpha_{\min}}
			=
			\frac{d}{d\alpha}s(\alpha).
		\end{equation}
		Substituting the two slopes gives
		\begin{equation}
			-\frac{\gamma}{\alpha_{\min}}
			=
			-\frac{\alpha_{\max} - \alpha_{\min}}{v_\alpha},
		\end{equation}
		which yields
		\begin{equation}
			\gamma_{\mathrm{default}}
			=
			\frac{\alpha_{\min}(\alpha_{\max} - \alpha_{\min})}{v_\alpha}.
		\end{equation}
		
		With this default $\gamma$, our bit allocation method adapts to different \texttt{PL\_Alpha\_Hill} distributions. Models with concentrated \texttt{PL\_Alpha\_Hill} distributions require larger curvature to expose importance differences, while broader distributions require weaker amplification. Figure~\ref{fig:gamma_ablation} reports the sensitivity of AlphaQ to $\gamma$ across representative MoE models.
	}
	
	\subsection{Further Experiment Results}
	\label{app:further_main}
	As shown in Tables~\ref{tab:result_deepseekv2-lite}, \ref{tab:table-qwen}, and~\ref{table-mixtral}, we provide more detailed results across models and bit budgets. We compare AlphaQ against PMQ~\cite{huang2024mixture} and Uniform under four bits-per-layer budget settings: 2.0 / 2.5 / 3.0 / 3.5. For evaluation, we report perplexity (PPL$\downarrow$) on WikiText2 and average zero-shot accuracy (Avg.$\uparrow$) over six benchmarks: PIQA~\cite{piqa}, ARC-Easy, ARC-Challenge~\cite{arcEasyAndChallenge}, HellaSwag~\cite{hellaswag}, WinoGrande~\cite{winogrande}, and BoolQ~\cite{boolq} using the EleutherAI LM Harness~\cite{eval-harness}. 
	
\begin{figure}
		\centering
		\includegraphics[width=0.7\linewidth]{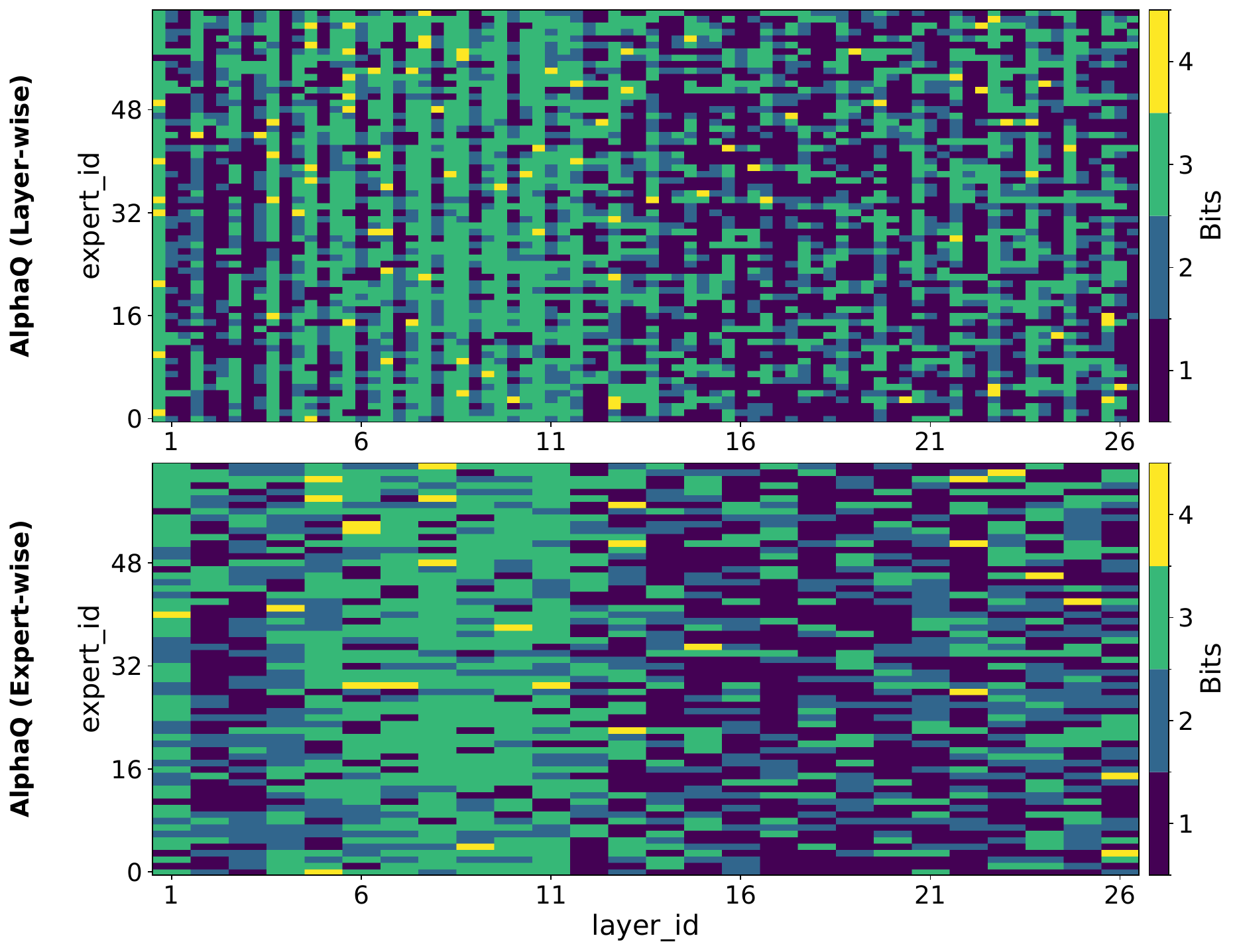}
		\caption{Bit allocation of DeepSeekV2-Lite under a 2-bit budget.}
		\label{fig:allocation_deepseek}
	\end{figure}

\begin{figure}
		\centering
		\includegraphics[width=0.7\linewidth]{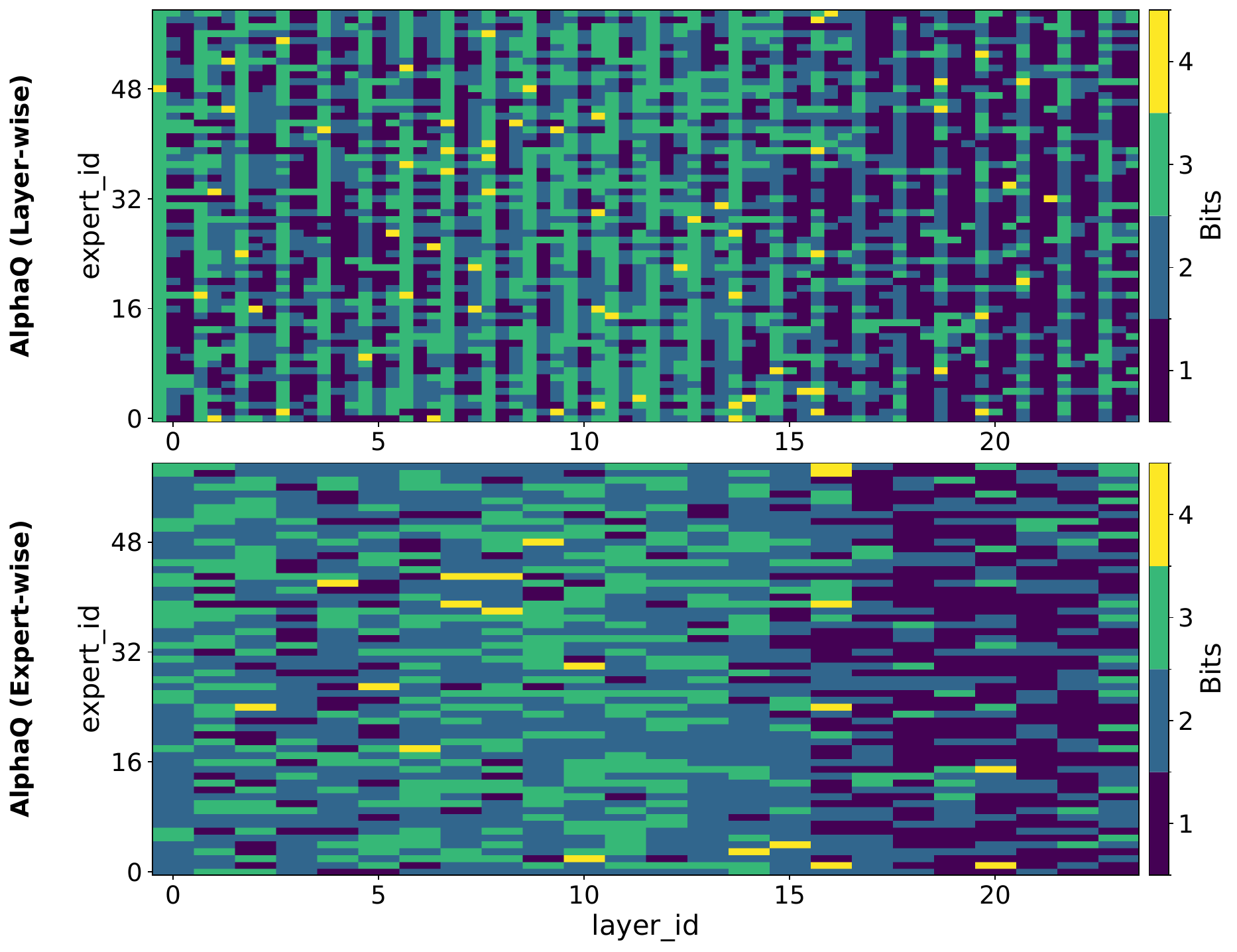}
		\caption{Bit allocation of Qwen1.5-MoE under a 2-bit budget.}
		\label{fig:allocation_qwen}
	\end{figure}

	In \cref{fig:allocation_deepseek}, \cref{fig:allocation_qwen}, and \cref{fig:allocation_mixtral}, we report detailed bit allocation results from AlphaQ under a 2-bit budget, with both layer-wise and expert-wise settings.

	
	\begin{table*}[htbp]
		\caption{Results on DeepSeekV2-Lite. Perplexity↓ on WikiText2 and accuracy↑ on six zero-shot tasks. The best results in each bit-width are highlighted in bold.}
		\label{tab:result_deepseekv2-lite}
		\centering
		\small
		\setlength{\tabcolsep}{4pt}
		\renewcommand{\arraystretch}{1.15}
		\begin{tabular}{ll c | cccccc c}
			\toprule
			\textbf{Bits} & \textbf{Method} & \textbf{WikiText2↓} & \textbf{PIQA↑} & \textbf{ARC-e↑} & \textbf{ARC-c↑} & \textbf{HellaSwag↑} & \textbf{WinoGrande↑} & \textbf{BoolQ↑} & \textbf{Avg.↑} \\
			\midrule
			16 & Original & 6.31 & 80.08 & 76.47 & 48.98 & 78.01 & 71.51 & 65.00 & 70.01 \\
			\midrule
			\multirow{3}{*}{2}
			& Uniform & 9.57 & 73.07 & 60.61 & 35.84 & 62.87 & 62.04 & 55.03 & 58.24 \\
			& PMQ & 7.99 & 75.14 & 61.68 & 37.37 & 66.20 & \textbf{68.11} & 57.66 & 61.03 \\
			& AlphaQ & \textbf{7.30} & \textbf{76.03} & \textbf{67.20} & \textbf{40.33} & \textbf{68.70} & 67.92 & \textbf{58.10} & \textbf{63.05} \\
			\midrule
			\multirow{3}{*}{2.5}
			& Uniform & 8.14 & 76.66 & 71.34 & 35.84 & 72.44 & 69.46 & 57.58 & 63.89 \\
			& PMQ & 6.95 & 77.86 & 71.68 & 37.37 & 72.76 & 69.06 & 60.03 & 64.79 \\
			& AlphaQ & \textbf{6.64} & \textbf{78.02} & \textbf{73.77} & \textbf{39.22} & \textbf{73.38} & \textbf{69.30} & \textbf{60.45} & \textbf{65.69} \\
			\midrule
			\multirow{3}{*}{3}
			& Uniform & 7.50 & 78.50 & 72.51 & 40.83 & 74.49 & 70.11 & 60.66 & 66.18 \\
			& PMQ & 6.70 & 79.16 & 73.15 & 41.44 & 75.04 & 70.23 & 61.53 & 66.76 \\
			& AlphaQ & \textbf{6.60} & \textbf{79.32} & \textbf{73.41} & \textbf{41.56} & \textbf{75.44} & \textbf{70.86} & \textbf{61.98} & \textbf{67.10} \\
			\midrule
			\multirow{3}{*}{3.5}
			& Uniform & 6.83 & 79.21 & 74.56 & 44.57 & 76.27 & 70.83 & 62.50 & 67.99 \\
			& PMQ & 6.57 & 79.63 & 75.26 & 45.83 & 76.97 & 71.04 & 63.21 & 68.66 \\
			& AlphaQ & \textbf{6.44} & \textbf{79.89} & \textbf{75.51} & \textbf{46.51} & \textbf{77.21} & \textbf{71.21} & \textbf{63.85} & \textbf{69.03} \\
			\bottomrule
		\end{tabular}
	\end{table*}
	
	\begin{table*}[t!]
		\caption{Results on Qwen1.5-MoE-A2.7B. Perplexity↓ on WikiText2 and accuracy↑ on six zero-shot tasks. The best results in each bit-width are highlighted in bold.}
		\label{tab:table-qwen}
		\centering
		\small
		\setlength{\tabcolsep}{4pt}
		\renewcommand{\arraystretch}{1.15}
		\begin{tabular}{ll c | cccccc c}
			\toprule
			\textbf{Bits} & \textbf{Method} & \textbf{WikiText2↓} & \textbf{PIQA↑} & \textbf{ARC-e↑} & \textbf{ARC-c↑} & \textbf{HellaSwag↑} & \textbf{WinoGrande↑} & \textbf{BoolQ↑} & \textbf{Avg.↑} \\
			\midrule
			16 & Original & 7.22 & 80.52 & 69.23 & 44.11 & 77.21 & 68.59 & 79.57 & 69.87 \\
			\midrule
			\multirow{3}{*}{2}
			& Uniform & 14.25 & 66.38 & 42.89 & 30.29 & 49.06 & 53.12 & 72.14 & 52.31 \\
			& PMQ & 10.47 & 74.76 & 59.05 & 34.56 & 64.34 & 64.72 & 74.52 & 61.99 \\
			& AlphaQ & \textbf{9.03} & \textbf{76.12} & \textbf{61.04} & \textbf{36.89} & \textbf{66.91} & \textbf{67.76} & \textbf{76.84} & \textbf{64.26} \\
			\midrule
			\multirow{3}{*}{2.5}
			& Uniform & 10.36 & 71.49 & 49.07 & 29.44 & 53.98 & 55.96 & 68.69 & 54.77 \\
			& PMQ & 9.01 & 77.58 & 60.56 & 36.26 & 70.69 & 65.82 & 75.54 & 64.41 \\
			& AlphaQ & \textbf{8.14} & \textbf{78.21} & \textbf{66.97} & \textbf{42.72} & \textbf{71.81} & \textbf{66.12} & \textbf{78.41} & \textbf{67.37} \\
			\midrule
			\multirow{3}{*}{3}
			& Uniform & 9.22 & 76.59 & 62.53 & 39.81 & 68.50 & 65.27 & 74.51 & 64.53 \\
			& PMQ & 7.85 & 79.82 & 68.19 & 42.55 & 76.15 & 68.22 & \textbf{77.89} & 68.80 \\
			& AlphaQ & \textbf{7.60} & \textbf{80.11} & \textbf{68.81} & \textbf{43.51} & \textbf{76.94} & \textbf{68.45} & 77.81 & \textbf{69.27} \\
			\midrule
			\multirow{3}{*}{3.5}
			& Uniform & 8.13 & 79.31 & 66.81 & 41.51 & 74.52 & 67.54 & 76.81 & 67.75 \\
			& PMQ & 7.42 & 80.31 & 69.02 & 43.81 & 77.13 & 68.50 & 79.22 & 69.67 \\
			& AlphaQ & \textbf{7.38} & \textbf{80.46} & \textbf{69.15} & \textbf{44.33} & \textbf{77.70} & \textbf{68.55} & \textbf{79.45} & \textbf{69.86} \\
			\bottomrule
		\end{tabular}
	\end{table*}

	\begin{table*}
		\caption{Results on Mixtral-8x7B. Perplexity↓ on WikiText2 and accuracy↑ on six zero-shot tasks. The best results in each bit-width are highlighted in bold.}
		\label{table-mixtral}
		\centering
		\small
		\setlength{\tabcolsep}{4pt}
		\renewcommand{\arraystretch}{1.15}
		\begin{tabular}{ll c | cccccc c}
			\toprule
			\textbf{Bits} & \textbf{Method} & \textbf{WikiText2↓} & \textbf{PIQA↑} & \textbf{ARC-e↑} & \textbf{ARC-c↑} & \textbf{HellaSwag↑} & \textbf{WinoGrande↑} & \textbf{BoolQ↑} & \textbf{Avg.↑} \\
			\midrule
			16 & Original & 3.84 & 83.57 & 83.67 & 59.30 & 84.03 & 76.24 & 85.35 & 78.69 \\
			\midrule
			\multirow{3}{*}{2}
			& Uniform & 6.30 & 74.81 & 71.09 & 40.44 & 54.04 & 65.75 & 72.57 & 63.12 \\
			& PMQ & 6.10 & 78.29 & 73.06 & 48.38 & \textbf{64.95} & 71.27 & \textbf{80.58} & \textbf{69.42} \\
			& AlphaQ & \textbf{6.01} & \textbf{78.52} & \textbf{74.62} & \textbf{46.59} & 64.76 & \textbf{71.31} & 80.39 & 69.36 \\
			\midrule
			\multirow{3}{*}{2.5}
			& Uniform & 6.16 & 79.38 & 77.40 & 50.26 & 58.84 & 73.40 & 82.01 & 70.22 \\
			& PMQ & 5.32 & 80.36 & 78.45 & 51.58 & 69.23 & \textbf{73.95} & 82.54 & 72.35 \\
			& AlphaQ & \textbf{5.17} & \textbf{80.84} & \textbf{78.87} & \textbf{51.59} & \textbf{69.24} & 73.24 & \textbf{82.57} & \textbf{72.39} \\
			\midrule
			\multirow{3}{*}{3}
			& Uniform & 4.91 & 80.20 & 79.55 & 49.57 & 70.23 & 74.19 & 82.01 & 72.62 \\
			& PMQ & 4.41 & 81.39 & \textbf{82.53} & 53.75 & 73.03 & 75.77 & 83.18 & 74.94 \\
			& AlphaQ & \textbf{4.37} & \textbf{81.56} & 82.32 & \textbf{54.44} & \textbf{73.47} & \textbf{77.27} & \textbf{83.64} & \textbf{75.45} \\
			\midrule
			\multirow{3}{*}{3.5}
			& Uniform & 4.29 & 81.28 & 83.04 & 54.95 & 73.58 & \textbf{76.19} & 84.36 & 75.57 \\
			& PMQ & 4.25 & 81.21 & 82.49 & 53.92 & 72.90 & 76.09 & 83.84 & 75.08 \\
			& AlphaQ & \textbf{4.23} & \textbf{81.51} & \textbf{83.38} & \textbf{55.80} & \textbf{74.67} & 76.10 & \textbf{84.45} & \textbf{75.98} \\
			\bottomrule
		\end{tabular}
	\end{table*}
	
	{\newupdate
		\begin{figure}
			\centering
			\includegraphics[width=0.7\linewidth]{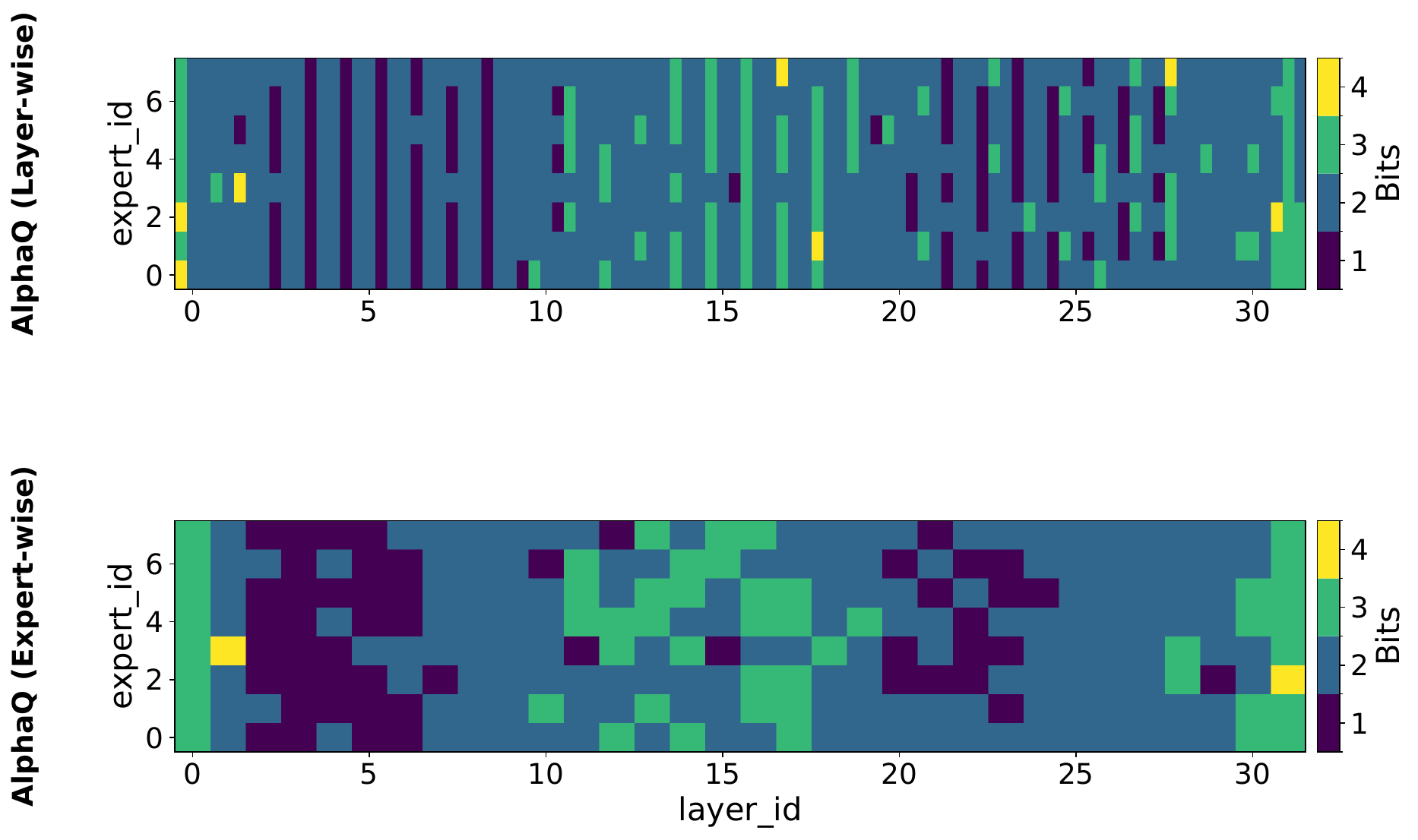}
			\caption{Bit allocation of Mixtral-8$\times$7B under a 2-bit budget.}
			\label{fig:allocation_mixtral}
		\end{figure}
		\subsection{Comparison with DynaMo}
		\label{app:dynamo_compare}
		\begin{table}[t]
			\centering
			\caption{\textbf{Comparison with DynaMo on OLMoE-1B-7B at 3-bit.} DynaMo numbers are taken from the reported results in original paper~\cite{zheng2025dynamo}. Lower is better for WikiText/C4 PPL, and higher is better for downstream accuracy.}
			\label{tab:dynamo_compare}
			\resizebox{0.85\linewidth}{!}{
				\begin{tabular}{l c c c c c}
					\toprule
					\textbf{Method} & \textbf{WikiText} & \textbf{C4} & \textbf{ARC-Challenge} & \textbf{ARC-Easy} & \textbf{PIQA} \\
					\midrule
					DynaMo & 9.64 & 15.31 & 25.85 & 43.52 & 58.87 \\
					AlphaQ & \textbf{9.19} & \textbf{14.64} & \textbf{26.91} & \textbf{46.11} & \textbf{60.08} \\
					\bottomrule
			\end{tabular}}
		\end{table}
	}
	
	{\newupdate
		For reference, we also compare AlphaQ with the recent cross-dataset MoE quantization method DynaMo~\cite{zheng2025dynamo}. Table~\ref{tab:dynamo_compare} shows that AlphaQ performs better than DynaMo on the overlapping OLMoE benchmarks. A possible reason is that DynaMo relies on a heuristic token-utilization signal to define channel and expert importance and evaluates cross-dataset adaptation only on two general-purpose corpora (WikiText and C4). By contrast, AlphaQ avoids calibration-driven importance estimation altogether and still transfers better under the same OLMoE 3-bit setting.
	}
	
	{\newupdate
		\subsection{Speedup and Memory Compression}
		\label{app:acc-speed}
		\begin{table}[t]
			\centering
			\caption{\textbf{Efficiency analysis on Mixtral-8$\times$7B and Qwen1.5-MoE.} 
				We report speedup relative to the BF16 baseline, total model size (Params), 
				and activated parameters during inference.}
			\label{tab:efficiency_table}
			\resizebox{0.95\linewidth}{!}{
				\begin{tabular}{l l c c c c}
					\toprule
					\textbf{Model} & \textbf{Method} & \textbf{Budget} & \textbf{Speedup $\uparrow$} & \textbf{Params (GB) $\downarrow$} & \textbf{Act. Params (GB) $\downarrow$} \\
					\midrule
					\multirow{8}{*}{\textbf{Mixtral-8$\times$7B}}
					& BF16    & 16-bit  & 1.00$\times$ & 96.80 & 26.31 \\
					& Uniform & 2-bit   & 1.70$\times$ & 13.61 & 3.70 \\
					& PMQ     & 2-bit   & 1.66$\times$ & 13.41 & 3.73 \\
					& PMQ     & 2.5-bit & 1.47$\times$ & 16.24 & 4.53 \\
					& AlphaQ  & 2-bit   & 1.69$\times$ & 13.41 & 3.75 \\
					& AlphaQ  & 2.5-bit & 1.50$\times$ & 16.30 & 4.52 \\
					& AlphaQ  & 3-bit   & 1.31$\times$ & 19.12 & 5.27 \\
					& AlphaQ  & 3.5-bit & 1.17$\times$ & 22.03 & 5.81 \\
					\midrule
					\multirow{3}{*}{\textbf{Qwen1.5-MoE}}
					& BF16   & 16-bit  & 1.00$\times$ & 28.94 & 5.74 \\
					& AlphaQ & 2.5-bit & 1.22$\times$ & 6.37 & 1.16 \\
					& AlphaQ & 3.5-bit & 1.12$\times$ & 7.60 & 1.45 \\
					\bottomrule
			\end{tabular}}
		\end{table}
		For memory compression, as shown in~\cref{tab:efficiency_table}, AlphaQ effectively addresses the memory bottleneck of MoE models. On Mixtral 8$\times$7B, the parameter memory footprint is reduced from 96.8~GB (BF16) to approximately 13.4~GB under a 2-bit budget ($7.2\times$ compression ratio). Under a 3.5-bit budget, AlphaQ achieves a $4.4\times$ compression ratio. With our optimized kernel, AlphaQ keeps a similar memory footprint to PMQ under comparable low-bit budgets, while the measured speedup is slightly higher at both 2-bit and 2.5-bit. Notably, under a 3.5-bit budget, Qwen1.5-MoE reduces the weight memory footprint to approximately one quarter of BF16.
		
	}
	
	\subsection{Runtime Implementation and Decode Optimization}
	\label{app:decode_runtime}
	We use an optimized runtime backend for MoE expert layers. During prefill, we reuse the PMQ-style HQQ~\cite{badri2023hqq} backend: a CUDA kernel performs bit unpacking and dequantization, followed by Tensor Core GEMM. Profiling shows this decomposition is highly effective in prefill, which is compute-bound, but yields limited speedups in memory-bound decode due to data movement between the dequantization and GEMM; concretely, \texttt{Memcpy HtoD} and \texttt{aten::copy\_} become dominant in decode. We therefore implement decode-specific fused Triton kernels that fuse unpacking, dequantization, and GEMM into a single kernel. Profiling confirms that this change substantially reduces \texttt{Memcpy HtoD} and \texttt{aten::copy\_} overhead during decode.
	
	Optionally, on top of this backend, we cache per-layer quantization metadata on the GPU during token generation, avoiding repeated host-to-device transfers across decode steps. This caching introduces a modest peak-memory increase: under the same WikiText2 benchmark on Mixtral-8$\times$7B, peak GPU memory rises from 13418MiB to 14259MiB (+841MiB, $\sim$6.1\%) for the 2-bit quantized model, which remains acceptable for most GPUs. 
	
	Table~\ref{tab:decode_latency_opt} reports end-to-end speedups on WikiText alongside decode token throughput. Relative to the PMQ-style backend, the fused kernel improves the decode token rate by nearly 1.5$\times$; adding metadata caching further improves throughput to 10.3 tokens/s.

	\begin{table}[t]
		\centering
		\caption{\textbf{WikiText2 decode performance with different backends on Mixtral-8$\times$7B.} The reported gain is measured on the same 2-bit-quantized model.}
		\label{tab:decode_latency_opt}
		\resizebox{0.7\linewidth}{!}{
			\begin{tabular}{l c c c c c}
				\toprule
				\textbf{Decode Backend} & \textbf{Token Rate (token/sec)$\uparrow$} \\
				\midrule
				\texttt{Unpack + Dequant + GEMM}    & 6.43 \\
				\texttt{Fused Kernel}  & 9.84  \\
				\texttt{Fused Kernel + metadata caching} & 10.36 \\
				\bottomrule
		\end{tabular}}
	\end{table}

	\subsection{Limitation}
	This work still has several limitations. 
	
	First, our experiments are restricted to weight-only quantization. Extending AlphaQ to activation bit allocation requires broader validation; we leave this to future work.

    	Second, AlphaQ relies on the degree of heavy-tailedness in the weight spectrum as a proxy for expert- and layer-level importance. While this assumption holds for the MoE language models evaluated in this work, and it is also known to hold for computer vision models~\cite{MM20a_trends_NatComm}, its generality for non-language models or broader MoE architectures remains to be validated. 
	
	Third, we focus solely on reducing domain bias in bit allocation. In quantization, data-dependent procedures that are orthogonal to allocation, such as error compensation in GPTQ, can still inject calibration-induced domain bias; we leave that to future work.
	
	Finally, we do not provide direct comparisons with all existing MoE quantization and compression methods, which may limit a comprehensive assessment of the method’s relative advantages. We leave these directions for future work and plan to further extend AlphaQ to broader MoE architectures and more comprehensive empirical studies.
	
	
	
\end{document}